\documentclass[lettersize,journal]{IEEEtran}

\usepackage[numbers]{natbib}
\usepackage{amsfonts}
\usepackage{mathtools}
\usepackage{hyperref}  
\usepackage{multirow}
\usepackage{booktabs} 
\usepackage{graphicx}
\usepackage{tikz}  
\usepackage{xparse}
\usepackage{lipsum}  
\usepackage{wasysym}

\usepackage{tabularx}

\newcolumntype{R}{>{\raggedleft\arraybackslash}X}
\newcolumntype{T}{>{\centering\arraybackslash}X}

\usepackage[indentafter]{parnotes}

\usepackage[nolist,nohyperlinks]{acronym}  
\usepackage{makecell} 
 
\usepackage{algorithm}
\usepackage[noend]{algpseudocode}

\DeclareMathOperator*{\EX}{\mathop{\mathbb{E}}}

\DeclareMathOperator{\bp}{\mathbf{p}}
\DeclareMathOperator{\bq}{\mathbf{q}}
\DeclareMathOperator{\x}{\mathbf{x}}

\DeclareMathOperator{\z}{\mathbf{z}}
\DeclareMathOperator{\p}{\mathbb{P}}
\DeclarePairedDelimiter{\norm}{\lVert}{\rVert}

\newcommand*\emptycirc[1][0.8ex]{\tikz\draw (0,0) circle (#1);} 
\newcommand*\halfcirc[1][0.8ex]{
  \begin{tikzpicture}
  \draw[fill] (0,0)-- (90:#1) arc (90:270:#1) -- cycle ;
  \draw (0,0) circle (#1);
  \end{tikzpicture}}
\newcommand*\fullcirc[1][0.8ex]{\tikz\fill (0,0) circle (#1);} 

\begin{document}

\title{ARCADE: Adversarially Regularized Convolutional Autoencoder for Network Anomaly Detection}

\author{Willian\,T.\,Lunardi,~\IEEEmembership{Member,~IEEE,}~Martin\,Andreoni Lopez,~\IEEEmembership{Member,~IEEE}~and~Jean-Pierre\,Giacalone%

\IEEEcompsocitemizethanks{\IEEEcompsocthanksitem Willian T. Lunardi, Martin Andreoni Lopez and Jean-Pierre Giacalone are with the Secure System Research Center, Technology and Innovation Institute, Abu Dhabi, United Arab Emirates -- \texttt{\{willian, martin, jean-pierre\}@ssrc.tii.ae}
}}

\markboth{Preprint submitted to IEEE TNSM, Special Issue on ``Machine Learning and Artificial Intelligence'', April 30, 2022}%
{ } 

% \markboth{IEEE Transactions on Network and Service Management, Vol.~*, No.~*, MONTH~YYYY}%
% { }

\maketitle

\begin{abstract}
As the number of heterogenous IP-connected devices and traffic volume increase, so does the potential for security breaches. The undetected exploitation of these breaches can bring severe cybersecurity and privacy risks. Anomaly-based \acp{IDS} play an essential role in network security. In this paper, we present a practical unsupervised anomaly-based deep learning detection system called ARCADE (Adversarially Regularized Convolutional Autoencoder for unsupervised network anomaly DEtection). 
With a convolutional \ac{AE}, ARCADE automatically builds a profile of the normal traffic using a subset of raw bytes of a few initial packets of network flows so that potential network anomalies and intrusions can be efficiently detected before they cause more damage to the network. ARCADE is trained exclusively on normal traffic. An adversarial training strategy is proposed to regularize and decrease the \ac{AE}'s capabilities to reconstruct network flows that are out-of-the-normal distribution, thereby improving its anomaly detection capabilities. 
The proposed approach is more effective than state-of-the-art deep learning approaches for network anomaly detection. Even when examining only two initial packets of a network flow, ARCADE can effectively detect malware infection and network attacks. ARCADE presents 20 times fewer parameters than baselines, achieving significantly faster detection speed and reaction time.
\end{abstract}
 

\begin{acronym}
    \acro{SAE}{Stacked Autoencoder}
    \acro{DNN}{Deep Neural Network}
    \acro{DCGAN}{Deep Convolutional GAN}
    \acro{AE}{Autoencoder}
    \acro{VAE}{Variational Autoencoder}
    \acro{AAE}{Adversarial Autoencoder}
    \acro{GAN}{Generative Adversarial Networks}
    \acro{DL}{Deep Learning}
    \acro{CNN}{Convolutional Neural Network} 
    \acro{UAV}{Unmanned Aerial Vehicle}
    \acro{IoT}{Internet of the Things}
    \acro{RNN}{Recurrent Neural Network}
    \acro{DDoS}{Distributed Denial of Service}
    \acro{SGD}{Stochastic Gradient Descent}
    \acro{SSIM}{Structural Similarity Index Measure} 
    \acro{$k$-NN}{$k$-Nearest Neighbor}
    \acro{JSD}{Jensen-Shannon divergence}
    \acro{WGAN}{Wasserstein Generative Adversarial Networks}
    \acro{WGAN-GP}{Wasserstein Generative Adversarial Networks with Gradient Penality}
    \acro{ML}{Machine Learning}
    \acro{LSTM}{Long Short-Term Memory}
    \acro{ReLU}{Rectified Linear Unit}
    \acro{Leaky ReLU}{Leaky Rectified Linear Unit}
    \acro{IDS}{Intrusion Detection System}
    \acro{FAR}{False Alarm Rate}
    \acro{DR}{Detection Rate}
    \acro{IP}{Internet Protocol}
    \acro{RNN}{Recurrent Neural Network}
    \acro{MLP}{Multilayer Perceptron}
    \acro{VPN}{Virtual Private Network}
    \acro{ANN}{Artificial Neural Network}
    \acro{BF}{Brute Force}
    \acro{PCA}{Principal Component Analysis}
    \acro{GRU}{Gated Recurrent Units}
    \acro{RNN}{Recurrent Neural Network}
\end{acronym}

\section{Introduction}

\IEEEPARstart{T}{he} proliferation of IP-connected devices is skyrocketing and is predicted to surpass three times the world's population by 2023~\cite{cisco2020cisco}. As the number of connected devices increases and 5G technologies become more ubiquitous and efficient, network traffic volume will follow suit. This accelerated growth raises overwhelming security concerns due to the exchange of vast amounts of sensitive information through resource-constrained devices and over untrusted heterogeneous technologies and communication protocols. Advanced security controls and analysis must be applied to maintain a sustainable, reliable, and secure cyberspace. \acp{IDS} play an essential role in network security, allowing for detecting and responding to potential intrusions and suspicious activities by monitoring network traffic. \acp{IDS} can be implemented as signature-based, anomaly-based, or hybrid. Signature-based \acp{IDS} detect intrusions by comparing monitored behaviors with pre-defined intrusion patterns, while anomaly-based \acp{IDS} focus on knowing normal behavior to identify any deviation~\cite{liao2013intrusion}.

The vast majority of existing network \acp{IDS} are based on the assumption that traffic signatures from known attacks can be gathered so that new traffic can be compared to these signatures for detection. Despite high detection capabilities for known attacks, signature-based approaches cannot detect novel attacks since they can only detect attacks for which a signature was previously created. Regular database maintenance cycles must be performed to add novel signatures for threats as they are discovered. Acquiring labeled malicious samples, however, can be extremely difficult or impossible to obtain. The definition of signature-based \acp{IDS}, or any other supervised approach for the task, becomes even more challenging when the known class imbalance problem is faced while dealing with public network traffic datasets is considered. Network traffic datasets are known for being highly imbalanced towards examples of normality (non-anomalous/non-malicious)~\cite{silva2022statistical} while lacking in examples of abnormality (anomalous/malicious) and offering only partial coverage of all possibilities can encompass this latter class~\cite{ahmad2021network}. 

In contrast, anomaly-based \acp{IDS} relies on building a profile of the normal traffic. These systems attempt to estimate the expected behavior of the network to be protected and generate anomaly alerts whenever a divergence between a given observation and the known normality distribution exceeds a pre-defined threshold. Anomaly-based \acp{IDS} do not require a recurrent update of databases to detect novel attack variants, and their main drawback usually is the \ac{FAR}, as it is challenging to find the boundary between the normal and abnormal profiles. These approaches have gained popularity in recent years due to the explosion of attack variants~\cite{truong2020empirical,hwang2020unsupervised}, which relates to their ability to detect previously unknown or zero-day threats. Additionally, they do not suffer from the dataset imbalance problem since it only requires normal traffic during training.

\ac{DL} has emerged as a game-changer to help automatically build network profiles through feature learning. It can effectively learn structured and complex non-linear traffic feature representations directly from the raw bytes of a large volume of normal data. Based on a well-represented traffic profile, it is expected that the system's capabilities for isolating anomalies from the normal traffic will increase while decreasing the \ac{FAR}. However, the naive adoption of \ac{DL} may lead to misleading design choices and the introduction of several drawbacks, such as slow detection and reaction time. In addition to carefully defining the model's architecture, training artifices could be exploited to improve the method's effectiveness without degrading the efficiency due to the increased number of parameters and model size.

In this paper, we propose ARCADE, an unsupervised \ac{DL} approach for network anomaly detection that automatically builds a profile of the normal traffic (training exclusively on the normal traffic) using a subset of bytes of a few initial packets of network traffic flow as input data. 
It allows early attack detection preventing any further damage to the network security while mitigating any unforeseen downtime and interruption. The network traffic can be originated from real-time packet sniffing over a network interface card or from a \texttt{.pcap} file.
The proposed approach combines two deep neural networks during training: (i)~an \ac{AE} trained to encode and decode (reconstruct) normal traffic; (ii)~a critic trained to provide high score values for real normal traffic samples, and low scores values for their reconstructions. An adversarial training strategy is settled where the critic’s knowledge regarding the normal traffic distribution is used to regularize the \ac{AE}, decreasing its potential to reconstruct anomalies, addressing the known generalization problem~\cite{rudolph2021same, gong2019memorizing, zhai2016deep}, where (in some scenarios) anomalies are reconstructed as well as normal samples. During detection, the error between the input traffic sample and its reconstruction is used as an anomaly score, i.e., traffic samples with high reconstruction error are considered more likely to be anomalous. The significant contributions of this paper are summarized as follows:
\begin{itemize}
    %\item An unsupervised DL-based approach for early anomaly detection that automatically builds the network traffic profile based on 100 raw bytes of $n \in \{1,2,3,4,5,6\}$ initial packet network flows of the normal traffic. It can detect (novel) network traffic anomalies given a few initial packets of network flows, allowing it to prevent network attacks before they could cause further damage.
    \item An unsupervised DL-based approach for early anomaly detection that automatically builds the network traffic profile based on the raw packet bytes of network flows of the normal traffic. It can detect (novel) network anomalies given a few initial packets of network flows, allowing it to prevent network attacks before they could cause further damage.

    \item A \ac{WGAN-GP} adversarial training to regularize \acp{AE}, decrease its generalization capabilities towards out-of-the-normal distribution samples, and improve its anomaly detection capabilities. 
    
    \item A compact convolutional~\ac{AE} model inspired by~\ac{DCGAN}~\cite{radford2015unsupervised}. The model presents higher accuracy, faster reaction time, 20 times fewer parameters than baselines.
    
    %\item A convolutional~\ac{AE} that is suitable for online resource-constrained environments. The model presents 20 times fewer parameters than baselines, achieving considerably higher accuracy and smaller reaction time when compared to baselines.
    
    \item An extensive validation of ARCADE conducted on several network traffic datasets to assess its capabilities in detecting anomalous traffic of several types of malware and attacks.
\end{itemize}

The remainder of the paper is laid out as follows: Section~\ref{sec:background} provides the necessary background for \acp{GAN}. Section~\ref{sec:rl} reviews and discusses previous relevant works in the field of \ac{DL} for anomaly detection and network anomaly detection. Section~\ref{sec:arcade} describes the proposed network flows preprocessing pipeline, model architecture, loss functions, and adversarial training strategy. Section~\ref{sec:experiments} presents the ablation studies and experimental analysis and comparison of ARCADE's effectiveness and complexity with respect to the considered baselines. Finally, Section~\ref{sec:conclusion} concludes this paper. 
 
\section{Background}\label{sec:background}
\subsection{Generative Adversarial Networks}\label{sec:gan}
The \acp{GAN}~\cite{goodfellow2014generative} framework establishes a min-max adversarial game between a generative model $G$ and a discriminative model $D$. The discriminator $D(\x)$ computes the probability that a point $\x$ in data space is a sample from the data distribution rather than a sample from our generative model. The generator $G(\z)$ maps samples $\z$ from the prior $p(\z)$ to the data space. $G(\z)$ is trained to maximally confuse the discriminator into believing that the samples it generates come from the data distribution. The process is iterated, leading to the famous minimax game~\cite{goodfellow2014generative} between generator $G$ and discriminator $D$
\begin{equation}
    \min_G\max_D \quad \EX_{x \sim \p_r} \log\big(D(\x)\big) + \EX_{\tilde x \sim \p_g} \log\big(1-D(\tilde\x)\big),
\end{equation} 
where $\p_r$ is the data distribution and $\p_g$ is the model distribution implicitly defined by $\tilde\x = G(\z)$, where $\z \sim p(\z)$ is the noise drawn from an arbitrary prior distribution.

Suppose the discriminator is trained to optimality before each generator parameter update. In that case, minimizing the value function amounts to minimizing the \ac{JSD} between $\p_r$ and $\p_g$~\cite{goodfellow2014generative}, but doing so often leads to vanishing gradients as the discriminator saturates~\cite{arjovsky2017wasserstein,gulrajani2017improved}.
 
\subsection{Wasserstein Generative Adversarial Networks}\label{sec:wgan}
To overcome the undesirable \ac{JSD} behavior, \citet{arjovsky2017wasserstein} proposed \ac{WGAN} that leverages Wasserstein distance $W(q,p)$ to produce a value function that has better theoretical properties than the original. They modified the discriminator to emit an unconstrained real number (score) rather than a probability. In this context, the discriminator is now called a critic.
The min-max \ac{WGAN} training objective is given by 
\begin{equation}
    \min_G \max_{C} \quad \EX_{\x \sim \p_r} \big[C(\x)\big] - \EX_{\tilde\x \sim \p_g} \big[C(\tilde\x)\big].
\end{equation}
When the critic $C$ is Lipschitz smooth, this approach approximately minimizes the Wasserstein-1 distance $W(\p_r, \p_g)$. To enforce Lipschitz smoothness, the weights of $C$ are clipped to lie within a compact space $[-c, c]$. However, as described in~\cite{arjovsky2017wasserstein}, weight clipping is a terrible approach to enforcing the Lipschitz constraint.

\citet{gulrajani2017improved} proposed an alternative approach where a soft version of the constraint is enforced with a penalty on the gradient norm for random samples $\hat \x \sim \p_{\hat \x}$. When considering the \ac{WGAN-GP} proposed in~\cite{gulrajani2017improved}, the critic's loss is given by
\begin{equation}
    \EX_{\x \sim \p_r} \big[C(\x)\big] - \EX_{\tilde\x \sim \p_g} \big[C(\tilde\x)\big] + \lambda_{\text{C}} \mathcal{L}_{\text{GP}},
\end{equation}
where  $\lambda_{\text{C}}$ is the penalty coefficient, and
\begin{equation}\label{eq:gp}
    \mathcal{L}_{\text{GP}} = \EX_{\hat \x \sim \p_{\hat \x}} \big[(\norm{\nabla_{\hat \x} C(\hat \x)}_2 - 1)^2 \big],
\end{equation}
where $\p_{\hat \x}$ is the distribution defined by the following sampling process: $\x \sim \p_r$, $\tilde\x \sim \p_g$, $\alpha \sim U(0, 1)$, and $\hat \x = \alpha \x + (1 - \alpha) \tilde\x$.

\begin{table*}[!t]
\setlength{\tabcolsep}{4pt}
\footnotesize	
\centering
\caption{Deep learning related works for network intrusion detection. For works that used raw network traffic as input, when specified, we present the number of packets ($n$) and bytes ($l$) used as input.}\label{tab:rw}
    \begin{tabularx}{\linewidth}{@{}>{\hsize=3.25cm}T>{\hsize=0.5cm}T>{\hsize=0.5cm}T>{\hsize=0.5cm}TT>{\hsize=5cm}T>{\hsize=5cm}T@{}} 
        \toprule
        \textbf{Paper} & \textbf{UD}\parnote{Unsupervised anomaly detection} & \textbf{AT}\parnote{Adversarial Training} & \textbf{RT}\parnote{Raw Traffic} & \textbf{Granularity} & \textbf{Input Data} & \textbf{Architecture} \\
        \midrule
        \citet{vu2017deep}                  & \emptycirc     &  \fullcirc    &  \emptycirc   & Session & Hand-designed features & Augmented dataset with AC-GAN \\
        \citet{truong2020empirical}         & \fullcirc      &  \fullcirc    &  \emptycirc   & Flow/Session       & Hand-designed features & MLP GAN with AE Generator \\
        \citet{doriguzzi2020lucid}          & \emptycirc    &  \emptycirc   &  \emptycirc   & Flow      & Hand-designed features & 1D-CNN  \\           
        \midrule
        \citet{wang2017hast}                & \emptycirc    &  \emptycirc   &  \fullcirc    & Flow & All layers, $l = [600, 800]$ & 2D-CNN and LSTM \\
        \citet{wang2017malware}             & \emptycirc    &  \emptycirc   &  \fullcirc    & Flow/Session & All/L7 layers, $l = 784$ & 2D-CNN \\
        \citet{yu2017network}               & \emptycirc    &  \emptycirc   &  \fullcirc    & Session & All layers, $l=1000$ & Dilated 2D-CNN \\
        \citet{wang2017end}                 & \emptycirc    &  \emptycirc   &  \fullcirc    & Flow/Session & All/L7 layers, $l=784$ & 1D-CNN \\  
        \citet{aceto2019mobile}             & \emptycirc     &  \emptycirc   &  \fullcirc   & Session & All/L7 layers, $n \in [4, 32]$, $l \in [256, 2304]$ & 1D/2D-CNN, LSTM \\        
        \citet{lotfollahi2020deep}          & \emptycirc    &  \emptycirc   &  \fullcirc    & Packet & IP packet, $l=1024$  & 2D-CNN \\ 
        \citet{hwang2020unsupervised}\parnote{Baseline}       & \halfcirc     &  \emptycirc   &  \fullcirc    & Flow & All layers, $n \in [2,5]$, $l \in [40, 80]$  & 1D-CNN with MLP AE \\        
        \citet{ahmad2022early}              & \emptycirc     &  \emptycirc   &  \fullcirc    & Session & All layers, $n\in[1,3]$, $ l = 450 $  & 1D-CNN \\        
        \midrule
        \textit{This paper}                 & \fullcirc     &  \fullcirc    &  \fullcirc    & Flow/Session & All layers, $n \in [2,5]$, $l = 100$ & 1D-CNN GAN with AE Generator \\
        \bottomrule
    \end{tabularx}
    \raggedright\scriptsize\parnotes
\end{table*}   
 
\section{Related Work}\label{sec:rl}
Herein, we discuss the relevant works employing \ac{DL} for anomaly detection. We first present DL anomaly detection approaches that have emerged as leading methodologies in the field of image and video. Then, we comprehensively analyze these novel DL methods and their potential application to network anomaly detection. We categorize unsupervised anomaly detection methods into generative models or pre-trained networks, introduced in Section~\ref{sec:generative} and~\ref{sec:pretrained}, respectively. Finally, Section~\ref{sec:network_ad_methods} presents the DL-related works for network traffic classification and baselines for unsupervised anomaly detection.

\subsection{Generative-based Anomaly Detection}\label{sec:generative}
Generative models, such as \acp{AE}~\cite{bergmann2018improving,hwang2020unsupervised} and \acp{GAN}~\cite{goodfellow2014generative,arjovsky2017wasserstein,gulrajani2017improved}, can generate samples from the manifold of the training data. Anomaly detection approaches using these models are based on the idea that anomalies cannot be generated since they do not exist in the training set.
 
\acp{AE} are neural networks that attempt to learn the identity function while having an intermediate representation of reduced dimension (or some sparsity regularization) serving as a bottleneck to induce the network to extract salient features from some dataset. These approaches aim to learn some low-dimensional feature representation space on which normal data instances can be well reconstructed. The heuristic for using these techniques in anomaly detection is that, since the model is trained only on normal data, normal instances are expected to be better reconstructed from the latent space than anomalies. Thus, the distance between the input data and its reconstruction can be used as an anomaly score. Although \acp{AE} have been successfully applied across many anomaly detection tasks, in some cases, they fail due to their strong generalization capabilities~\cite{rudolph2021same}, i.e., sometimes anomalies can be reconstructed as well as normal samples. \citet{bergmann2018improving} shows that \acp{AE} using the  \ac{SSIM}~\cite{wang2004image} can outperform complex architectures that rely on a per-pixel value $\ell_2$-loss. \citet{gong2019memorizing}~tackle the generalization problem by employing memory modules which can be seen as a discretized latent space. \citet{zhai2016deep}~connect regularized \acp{AE} with energy-based models to model the data distribution and classify samples with high energy as an anomaly.

\ac{GAN}-based approaches assume that only positive samples can be generated. These approaches generally aim to learn a latent feature space of a generative network so that the latent space well captures the normality underlying the given data~\cite{pang2021deep}. Some residual between the real and generated instances is then defined as an anomaly score. One of the early \ac{GAN}-based methods for anomaly detection is AnoGAN~\cite{schlegl2017unsupervised}. The fundamental intuition is that given any data instance $\x$; it aims to search for an instance $\z$ in the learned latent features space of the generative network $G$ so that the corresponding generated instance $G(\z)$ and $\x$ are as similar as possible. Since the latent space is enforced to capture the underlying distribution of training data, anomalies are expected to be less likely to have highly similar generated counterparts than normal instances. One main issue with AnoGAN is the computational inefficiency, which can be addressed by adding an extra network that learns the mapping from data instances onto latent space, i.e., an inverse of the generator, resulting in methods like EBGAN~\cite{zenati2018efficient}. \citet{akcay2018ganomaly} proposed GANomaly that further improves the generator over the previous works by changing the generator to an encoder-decoder-encoder network. The \ac{AE} is trained to minimize a per-pixel value loss, whereas the second encoder is trained to reconstruct the latent codes produced by the first encoder. The latent reconstruction error is used as an anomaly score. 

The idea behind \acp{AE} is straightforward and can be defined under different \ac{ANN} architectures. Several authors have already investigated the applicability of \acp{AE} for network anomaly detection~\cite{truong2020empirical,hwang2020unsupervised}. However, its naive adoption can lead to unsatisfactory performance due to its vulnerability to noise in the training data and its generalization capabilities. We propose an adversarial regularization strategy with a carefully designed and compact \ac{AE} parameterized by 1D-\ac{CNN}. The adversarial training is employed to deal with the aforementioned \ac{AE}'s weaknesses. Similarly to GANomaly, our approach employs an adversarial penalty term to the \ac{AE} to enforce it to produce normal-like samples. Therefore, we also consider the GANomaly framework as a baseline and compare it with the proposed ARCADE for network anomaly detection.

\subsection{Pretrained-based Anomaly Detection}\label{sec:pretrained}
Pretrained-based anomaly detection methods use models trained on large datasets, such as ImageNet, to extract features~\cite{deng2009imagenet}. These pre-trained models produce separable semantic embeddings and, as a result, enable the detection of anomalies by using simple scoring methods such as \ac{$k$-NN} or Gaussian Mixture Models~\cite{xiao2021we}. Surprisingly, the embeddings produced by these algorithms lead to good results even on datasets that are drastically different from the pretraining ones. Recently, \citet{bergman2020deep}~showed that using a \ac{$k$-NN} for anomaly detection as a scoring method on the extracted features of a pre-trained ResNet model trained on the ImageNet produces highly effective and general anomaly detection methods on images. That alone surpassed almost all unsupervised and self-supervised methods. In~\cite{reiss2021panda}, it is shown that fine-tuning the model using either center loss or contrasting learning leads to even better results.

The application of those methods for network anomaly detection is challenging primarily due to the detection's complexity related to the additional required scoring step. Even with a compact model, such as the proposed in Section~\ref{sec:architecture} with 184k parameters, or the EfficientNet B0 with 5.3M parameters, the requirement for a post-processing scoring procedure makes it unsuitable for online detection, e.g., after forwarding the sample through the model for feature extraction, computing the anomaly score for a given sample's feature vector with \ac{$k$-NN} as the scoring method (as proposed in~\cite{bergman2020deep}), implies $O(nl)$ time complexity, where $n$ is the number of training samples, and $l$ is the length of the feature vectors. These techniques appear unexplored and may stand out for \textit{offline} network anomaly detection.

\subsection{Deep Learning for Network Traffic Classification}\label{sec:network_ad_methods}
Several works have studied DL network traffic classification under the supervised setting. Few works also have studied adversarial training strategies for network traffic classification based on hand-designed features. Nonetheless, feature learning-based unsupervised network anomaly detection with adversarial training appears currently unexplored. Table~\ref{tab:rw} summarizes our related works, which are categorized into: (i)~Unsupervised anomaly detection (UD) when only normal traffic is considered at the training stage. (ii)~Adversarial Training (AT) when \ac{GAN}-based strategies are applied during training. (iii)~Raw Traffic (RT) when the input is the raw network traffic. When RT is considered, the considered protocol layers, the number of initial bytes $l$, and the number of packets $n$ are presented.

\begin{figure}[!t]
\centering
    \begin{tabular}{cccc}
        \includegraphics[width=0.2\columnwidth]{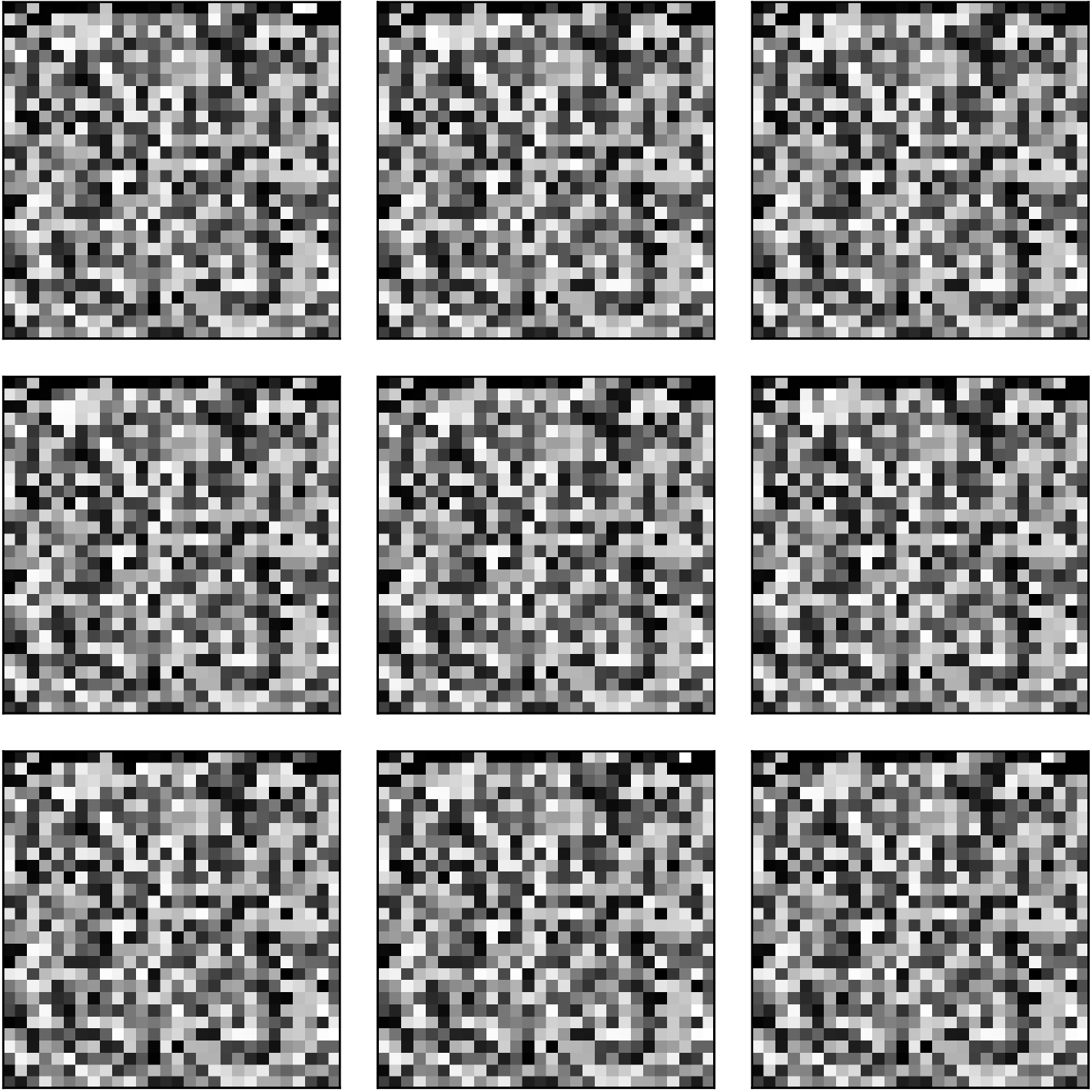}  &  \includegraphics[width=0.2\columnwidth]{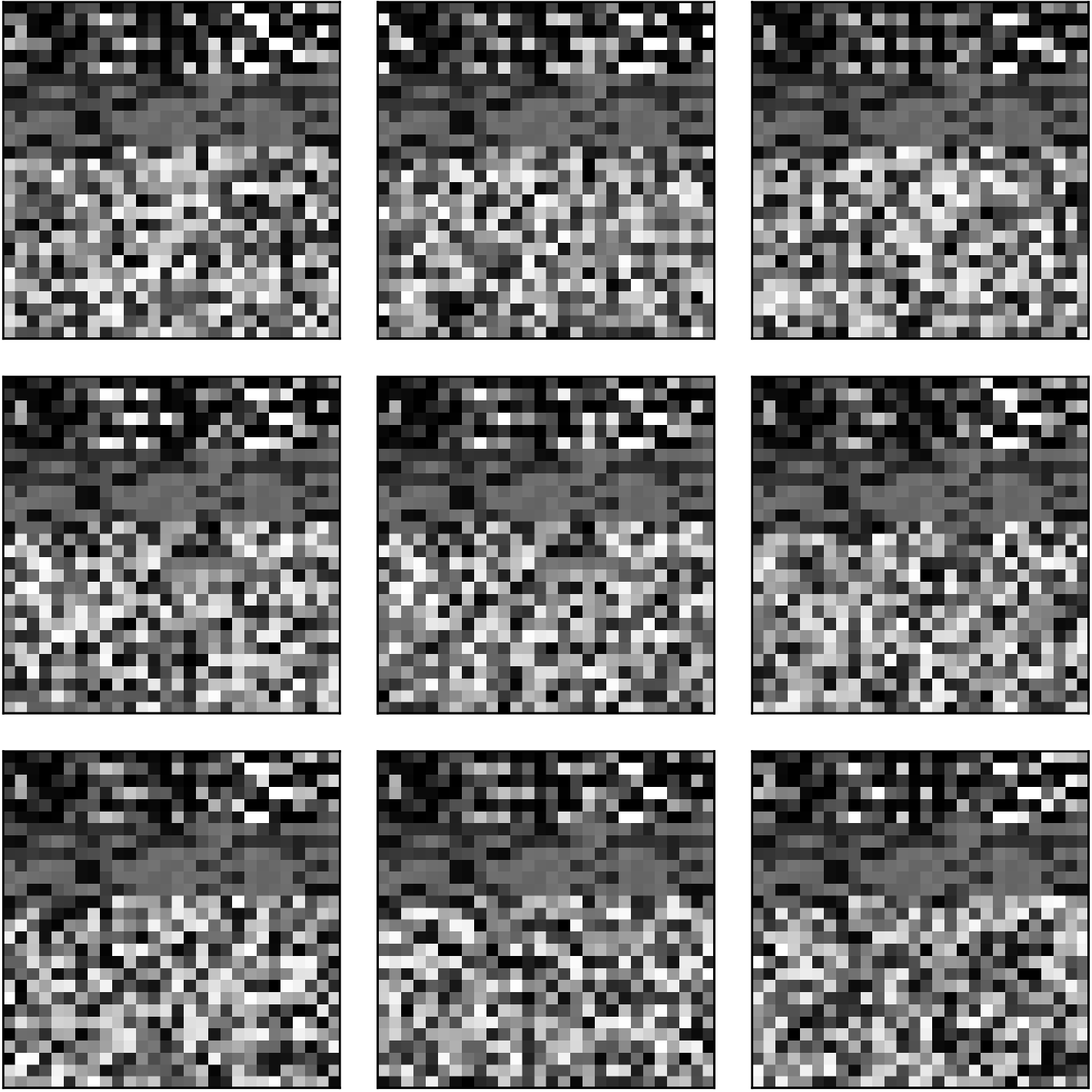}  & \includegraphics[width=0.2\columnwidth]{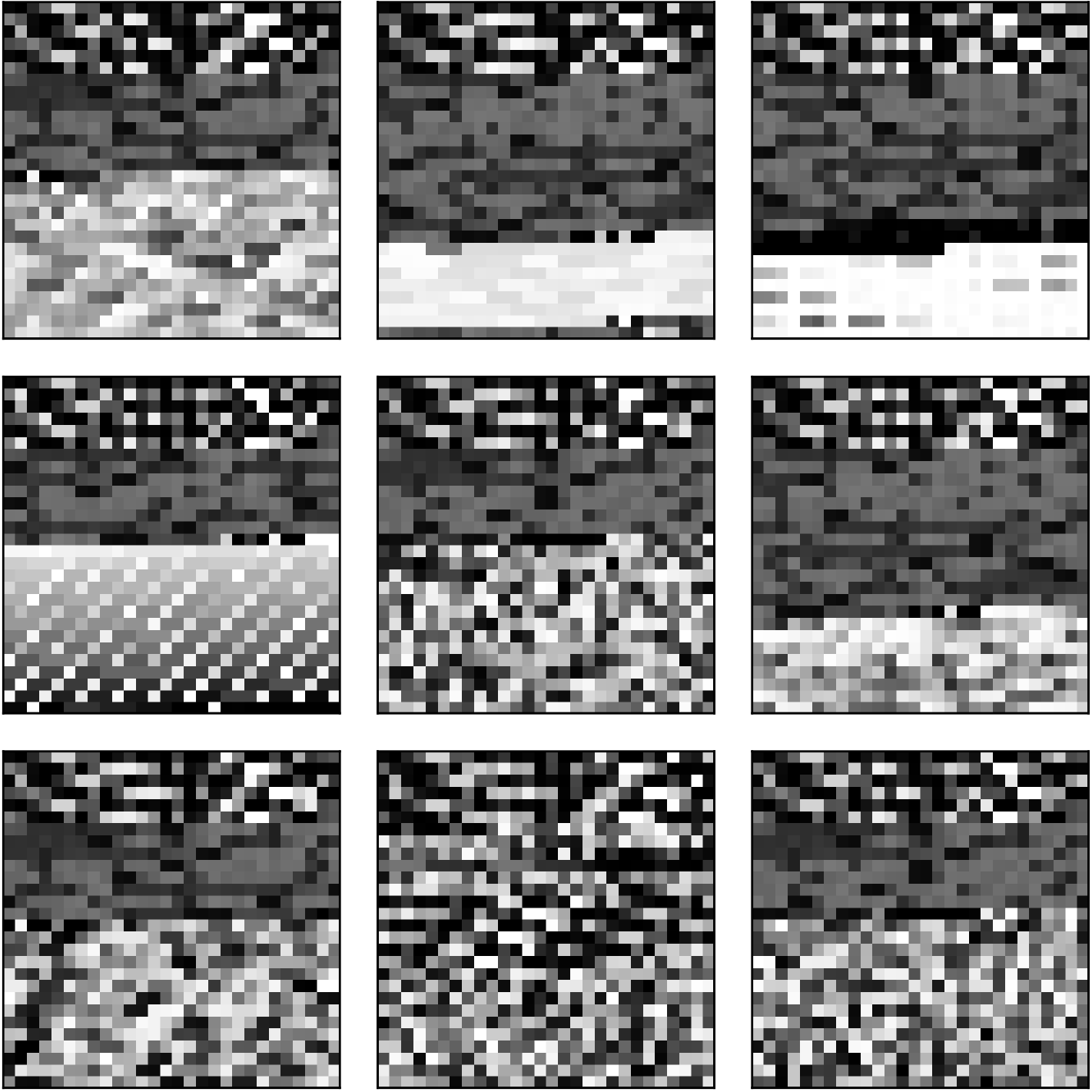} & 
        \includegraphics[width=0.2\columnwidth]{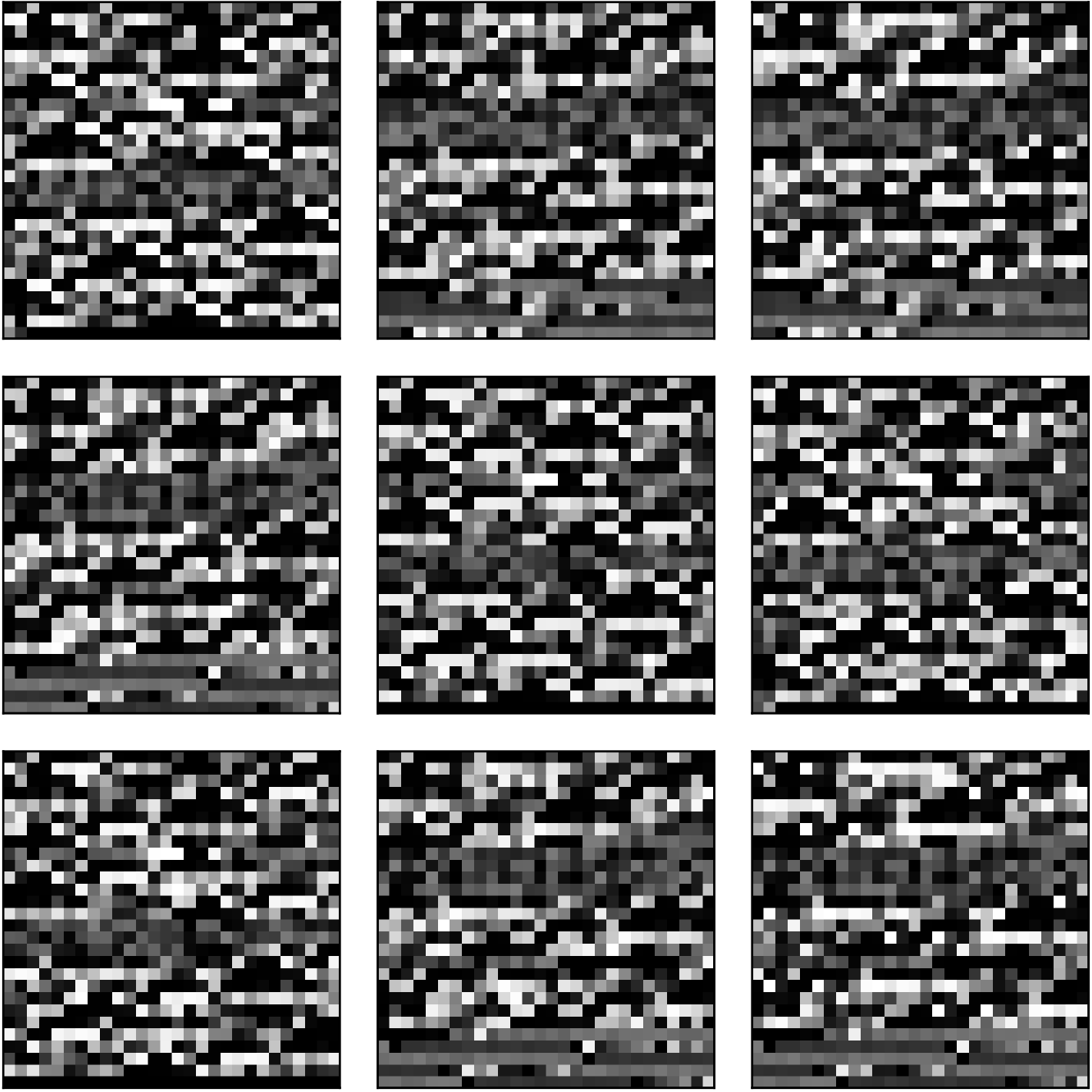}  \\
        (a) & (b) & (c) & (d) \\
    \end{tabular}
\caption{Visualization of network flows from four distinct traffic classes of the USTC-TFC dataset. In this instance, $784$ initial bytes of nine network flows (of four traffic classes) were reshaped into $28 \times 28$ grayscale images. (a) FTP. (b) Geodo. (c) Htbot. (d) World of Warcraft.}\label{fig:network_flows}
\end{figure} 

Most deep learning-based traffic classification and anomaly detection approaches rely on feature engineering. We highlight a few works in which adversarial training or unsupervised anomaly detection was addressed and which rely on hand-designed features. \citet{vu2017deep} proposed the use of a \acp{GAN} for dealing with the imbalanced data problem in network traffic classification. The synthetic samples are generated to augment the training dataset. The sample's generation, as well as the classification, is done based on 22 statistical features extracted from network flows. \citet{truong2020empirical} studied the capability of a \ac{GAN} for unsupervised network anomaly detection where 39 hand-designed features extracted from traffic flows and sessions are used as input. Results show that their proposed approach managed to obtain better results when compared to the autoencoder without any enhanced adversarial training. \citet{doriguzzi2020lucid} proposed a spatial representation that enables a convolutional neural network to learn the correlation between 11 packet's features to detect \ac{DDoS} traffic. 

Network traffic feature learning is predominantly performed through \ac{ANN} architectures like 1D-\ac{CNN}, 2D-\ac{CNN}, and \ac{LSTM}. Extracted bytes from network traffic flows (or packets) are kept sequential for the 1D-\ac{CNN} and \ac{LSTM} case, whereas for the 2D-\acp{CNN}, extracted bytes are seen as pixels of grayscale images, as illustrated in Figure~\ref{fig:network_flows}. \citet{wang2017hast} proposed an approach that relies on the advantages of both 2D-\acp{CNN} and \acp{LSTM} to extract spatial-temporal features of network traffic. Results show that accuracy is improved when both architectures are combined. \citet{wang2017malware} proposed a supervised \ac{DL} approach for malware traffic classification that uses 2D-\acp{CNN} to extract spatial features from headers and payloads of network flows and sessions. Two different choices of raw traffic images (named ``ALL" and ``L7") dependent on the protocol layers considered to extract the input data are used to feed the classifier, showing that sessions with ``ALL" are the most informative and reach elevate performance for all the metrics considered. \citet{yu2017network} proposed a self-supervised learning 2D-\ac{CNN} \ac{SAE} for feature extraction, which is evaluated through different classification tasks with malware traffic data. \citet{wang2017end} have shown that 1D-\ac{CNN} outperforms 2D-\ac{CNN} for encrypted traffic classification. \citet{aceto2019mobile} performed an in-depth comparison on the application of \ac{MLP}, 1D-\ac{CNN}, 2D-\ac{CNN}, and \ac{LSTM} architectures for encrypted mobile traffic classification. Numerical results indicated that 1D-\ac{CNN} is a more appropriate choice for network traffic classification since it can better capture spatial dependencies between adjacent bytes in the network packets due to the nature of the input data that is, by definition, one-dimensional. \citet{lotfollahi2020deep} used 1D-\ac{CNN} to automatically extract network traffic features and identify encrypted traffic to distinguish \ac{VPN} and non-\ac{VPN} traffic. \citet{ahmad2022early} employed 1D-\ac{CNN}-based classifier for early detection of network attacks. It is shown that a high degree of accuracy can be achieved by analyzing 1 to 3 packets.

The works mentioned above perform the task of traffic classification or anomaly detection based on labeled datasets. Recently, \citet{hwang2020unsupervised} proposed an ``unsupervised" approach for anomaly detection, so-called D-PACK, in which only normal traffic is used during training. The model architecture is composed of 1D-\ac{CNN} that performs feature extraction, followed by \ac{MLP} softmax classifier given a labeled dataset of normal traffic, i.e., they assume the normal traffic is labeled into multiple classes (that is the reason why its respective UD bullet is partially filled in Table~\ref{tab:rw}). The extracted features from an intermediate layer of the \ac{MLP} are used as the input for a \ac{MLP}-based \ac{AE}. The anomaly score is based on a $\ell_2$-distance between the extracted features and the \ac{AE} reconstruction. Results indicate that normal and malware traffic, such as the Mirai Botnet, can be effectively separated even when only two packets are used for detection. We implemented and included D-PACK in our experiments as a baseline model.
 
 \section{Methodology}\label{sec:arcade}
In this section, we present our so-called ``ARCADE" proposed approach. The network flow preprocessing procedure is presented in~\ref{sec:preprocessing}. The model's architecture is presented in Section~\ref{sec:architecture}. The \ac{AE} distance metrics and adversarial training are presented in Section~\ref{sec:ae_dist} and Section~\ref{sec:adv_training}, respectively. Finally, the anomaly score calculation is presented in Section~\ref{sec:anom_score}.

\begin{figure}[!t]
    \centering
    \includegraphics[width=\columnwidth]{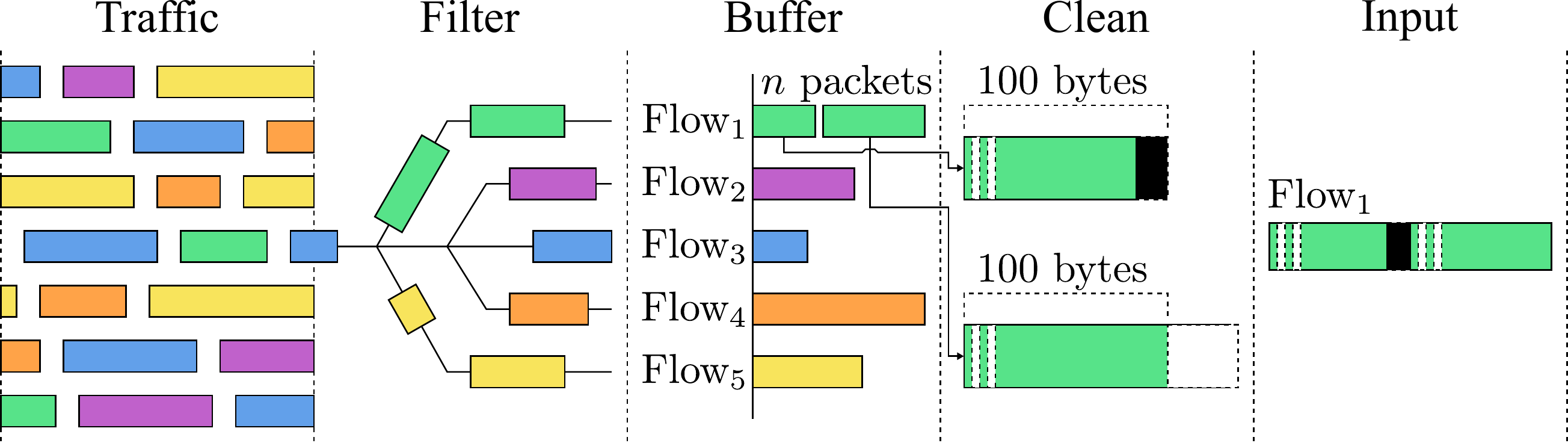}
    \caption{An illustration of the proposed network traffic preprocessing pipeline with $n = 2$. Packets with the same color represent network flows. Traffic can be originated from a real-time packet sniffing or a \texttt{.pcap} file. Packets are filtered according to their 5-tuple, and $n$ initial packets are buffered. MAC and IP addresses are masked, and according to their length, packets are truncated (if larger than $l$) or padded with zeros (if smaller than $l$). Finally, bytes are normalized, and packets are concatenated. Traffic preemption is not required.}\label{fig:preprocessing} 
\end{figure} 

\subsection{Network Traffic Flow Preprocessing}\label{sec:preprocessing}
Network traffic classification or anomaly detection can be performed at different granularity units, e.g., packet, flow, and session. It is worth noting that most of the works shown in Table~\ref{tab:rw} considered either flows or sessions as the relevant classification objects. A \textit{network flow} is a unidirectional sequence of packets with the same 5-tuple (source IP, source port, destination IP, destination port, and transport-level protocol) exchanged between two endpoints. A \textit{session} is defined as a bidirectional flow, including both directions of traffic. We increment the aforementioned flow definition by considering that a network flow is to be terminated or inactivated when the flow has not received a new packet within a specific flow timeout (e.g., 120 seconds). When the underlying network protocol is TCP, we consider the network connection closed (and the corresponding flow completed) upon detecting the first flow packet containing a FIN flag. Note that, in the case of TCP sessions, a network connection is considered closed when both sides have sent a FIN packet. Upon the termination of a network flow, unprocessed buffered packets should be discarded.

It is well known that the initial packets of each network flow contain the most information that allows for the discrimination between normal and abnormal activities~\cite{aceto2019mobile,ahmad2022early,hwang2020unsupervised}, depicting the fundamental concept behind early detection approaches, which conduct the detection given a small number of initial packets of a flow. The smaller the number of packets required as input for the anomaly detection procedure, the lower the reaction time and overhead imposed by the DL method. Instead of analyzing every packet of a network flow on a time window, we use the $n$ initial packets of a network flow as input. In this sense, $n$ denotes the exact number of initial packets of a network flow required to form the input for ARCADE. For each active flow, $n$ packets are buffered and trimmed into a fixed length of 100 bytes, starting with the header fields, i.e., packets are truncated to 100 bytes if larger, otherwise, padded with zeros. Packets are cleaned such that MAC and IP addresses are anonymized. Finally, bytes are normalized in $[0,1]$ and packets concatenated into the final input form, i.e., a sample $\x$ can be denoted as $\x \in \mathbb{R}^{w}$ where $w = 100n$ is the sequence length. Figure~\ref{fig:preprocessing} illustrates the essential steps of the proposed network traffic flow preprocessing pipeline. Note that traffic preemption is not required. However, it is crucial to consider the reaction time, which relates to the processing power capabilities of the device in which the ARCADE will run. The analysis of the complexity and detection speed of ARCADE given different devices is provided in Section~\ref{sec:complexity}.

\begin{figure}[!t]
    \centering
    \includegraphics[width=\columnwidth]{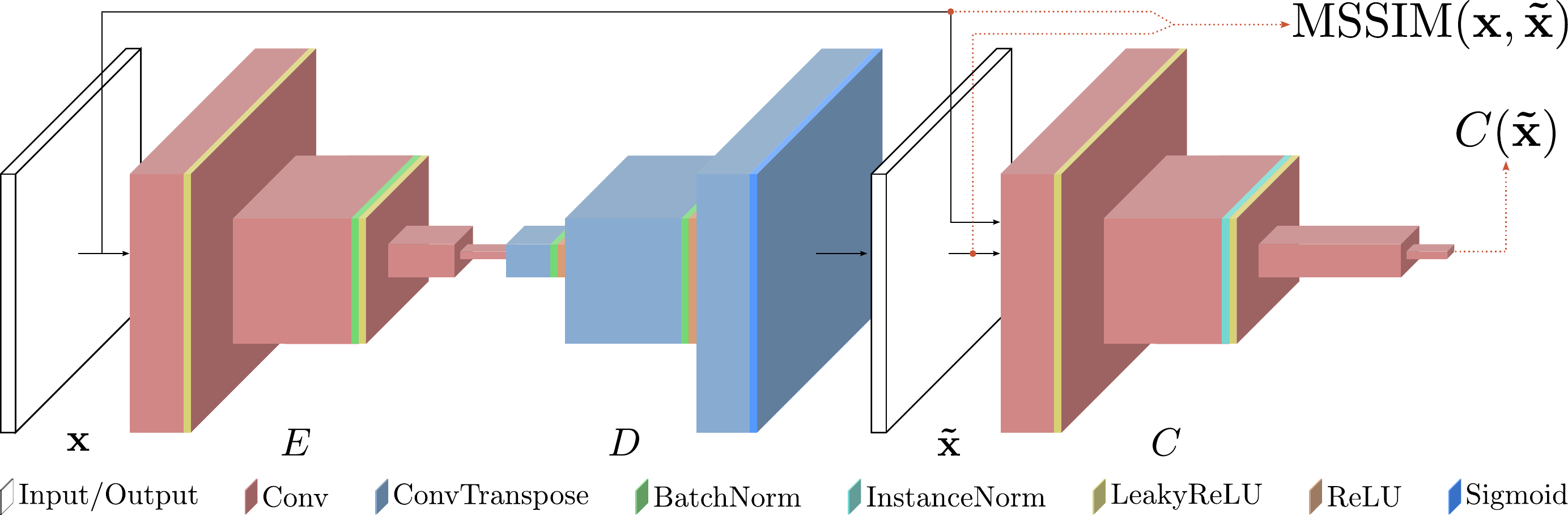} 
    \caption{An illustration of the proposed model architecture. Note that ARCADE is parametrized by 1D \acp{CNN}, as described in Section~\ref{sec:architecture}}\label{fig:architecture}  
\end{figure} 

\subsection{Model Architecture}\label{sec:architecture}
Several recent papers focus on improving training stability and the resulting quality of \ac{GAN} samples~\cite{radford2015unsupervised,arjovsky2017wasserstein,gulrajani2017improved}. Our proposed model is inspired by~\ac{DCGAN}~\cite{radford2015unsupervised}, who introduce a convolutional generator network by removing fully connected layers and using convolutional layers and batch-normalization throughout the network. Strided convolutions replace pooling layers. This results in a more robust model with higher sample quality while reducing the model size and number of parameters to learn.
Our proposed architecture shown in Figure~\ref{fig:architecture} consists of two main components: (i)~the \ac{AE} (which can be seen as the generator) composed of an encoder $E$ and a decoder $D$, and (ii)~the critic $C$. Given the findinds in~\cite{aceto2019mobile}, our functions $E$, $D$ and $C$ are parameterized by 1D-\acp{CNN}. Note that ARCADE can be easily modified to be used as an anomaly detection method for other anomaly detection tasks, such as image or time series anomaly detection. Moreover, the proposed adversarial regularization strategy can be applied to any \ac{AE}, independent of its \ac{ANN} architecture.

The \ac{AE} consists of an encoder function $E : \mathbb{R}^{w} \mapsto \mathbb{R}^d$ and a decoder function $D : \mathbb{R}^d \mapsto \mathbb{R}^{w}$, where $d$ denotes the dimensionality of the latent space. The overall encoding and decoding process can be summarized as 
\begin{equation}
    \tilde\x = D\big(E(\x)\big) = G(\x),
\end{equation}
where $\tilde\x$ is the reconstruction of the input. The encoder uses strided convolutions to down-sample the input, followed by batch normalization and \ac{Leaky ReLU}. Differently from a deterministic pooling operation, strided convolutions allow the model to learn its own downsampling/upsampling strategy. The decoder uses strided transpose convolutions to up-sample the latent space, followed by \ac{ReLU} and batch normalization. 
The critic function $C : \mathbb{R}^{w} \mapsto \mathbb{R}$, whose objective is to provide a score to the input $\x$ and the reconstruction $\tilde\x$, has a similar architecture to the encoder $E$. It also uses strided convolutions to down-sample the input and \ac{Leaky ReLU}; however, following~\cite{gulrajani2017improved}, we use layer normalization instead of batch normalization. The number of layers and filter size were defined so that ARCADE could effectively detect anomalies and still provide quick reaction time.
Table~\ref{tab:architecture} precisely presents the proposed model architecture.

\subsection{Autoencoder Distance Metric}\label{sec:ae_dist}
The core idea behind ARCADE is that the model must learn the normal traffic distribution to reconstruct it correctly. The hypothesis is that the model is conversely expected to fail to reconstruct attacks and malware traffic as it is never trained on such abnormal situations. A loss function must be defined to train an \ac{AE} to reconstruct its input. For simplicity, a per-value $\mathcal{L}_2$ loss is typically used between the input $\x$ and reconstruction $\tilde\x$, and can be expressed as
\begin{equation}
    \mathcal{L}_2(\x, \tilde\x) = \sum_{i=1}^w \big(\x_i - \tilde\x_i\big)^2,
\end{equation}
where $\x_i$ is the $i$-th value in the sequence. During the evaluation phase, the per-value $\ell_2$-distance of $\x$ and $\tilde\x$ is computed to obtain the residual map.

\begin{figure}[!t]
    \centering
    \includegraphics[width=\columnwidth]{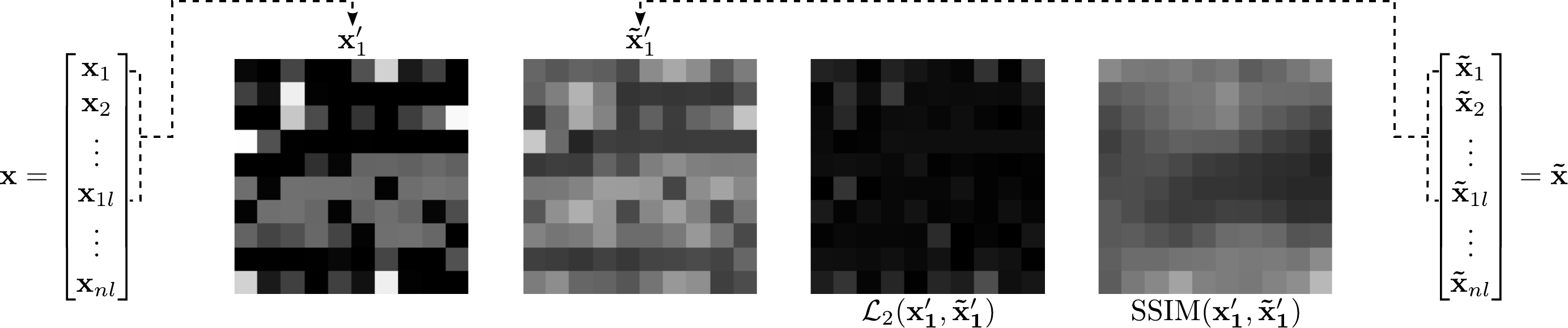} 
    \caption{An illustration of the advantages of the \ac{SSIM} over $\mathcal{L}_2$ for the segmentation of the discrepancies between a subset of bytes of a packet and their respective reconstructions.}\label{fig:ssim} 
\end{figure} 

As demonstrated by~\citet{bergmann2018improving}, \acp{AE} that make use of $\mathcal{L}_2$ loss may fail in some scenarios to detect structural differences between the input and their reconstruction. Adapting the loss and evaluation functions to the \ac{SSIM}~\cite{wang2004image} that capture local inter-dependencies between the input and reconstruction regions can improve the \ac{AE}'s anomaly detection capabilities. This is also verified in this work, as demonstrated in Section~\ref{sec:experiments}. The \ac{SSIM} index defines a distance measure between two $K \times K$ patches $\bp$ and $\bq$ is given by
\begin{equation}\label{eq:ssim}
    \text{SSIM}(\bp, \bq) = \frac{(2\mu_{\bp}\mu_{\bq} + c_1)(2\sigma_{\bp\bq} + c_2)}{(\mu_{\bp}^2 + \mu_{\bq}^2 + c_1)(\sigma_{\bq}^2 + \sigma_{\bq}^2 + c_2)},
\end{equation}
where $\mu_{\bp}$ and $\mu_{\bq}$ are the patches' mean intensity, $\sigma_{\bp}^{2}$ and $\sigma_{\bq}^{2}$ are the variances, and $\sigma_{\bp\bq}$ the covariance. The constants $c_1$ and $c_2$ ensure numerical stability and are typically set to $c_1 = 0.01$ and $c_2 = 0.03$.

The \ac{SSIM} is commonly used to measure the similarity between images, performed by sliding a $K \times K$ window that moves pixel-by-pixel. Since in our case $\x$ is a sequence, we split it into $n$ subsequences of length $l$, i.e., each subsequence 
\[
\x'_i = \big(\x_j \in [0,1] : j \in \{1 + (i - 1)l,\dots, il\}\big),
\]
where $i \in \{1, 2, \dots, n\}$ and $l=100$ can be seen as the subset of 100 bytes of the $i$-th packet that was originally used to compose the sequence $\x$. Finally, subsequences are reshaped $\x'_i \in \mathbb{R}^{l} \mapsto \x'_i \in \mathbb{R}^{K \times K}$, where $K = \sqrt{l}$ and $l$ is a perfect square number. An illustration of this procedure is shown in Figure~\ref{fig:ssim}. The mean SSIM gives the overall structural similarity measure of the sequence (MSSIM), defined as
\begin{equation}\label{eq:mssim}
    \text{MSSIM}(\x, \tilde\x) = \frac{1}{nM}\sum_{i=1}^n \sum\limits_{j=1}^{M}\text{SSIM}\big(\x'_i(j), \tilde\x'_i(j)\big),
\end{equation}
where $M$ is the number of local windows, and $\x'_i(j)$ and $\tilde\x'_i(j)$ are the contents at the $j$-th local window of the $i$-th subsequences $\x'_i$ and $\tilde\x'_i$. 

\begin{algorithm}[b!]
{\small
\caption{\small{Proposed adversarial training. We use $m=64$, $\lambda_{\text{C}} = 10$, $\lambda_{\text{G}} = 100$, $\alpha=\text{1e-4}$, $\beta_1 = 0$, and $\beta_2=0.9$.}}\label{pseudo:algo1}
\textbf{Require:} Batch size $m$, maximum training iterations $\max_{\text{epoch}}$, $C$ penalty coefficients $\lambda_{\text{C}}$ and $\lambda_{\text{G}}$, Adam hyperparameters $\alpha, \beta_1, \beta_2$, critic and autoencoder initial parameters $\psi_0$ and $\theta_0$, respectively.
\begin{algorithmic}[1]
    \While{current epoch is smaller than $\max_{\text{epoch}}$}
        \State Sample a batch of normal traffic samples $\{\x^{(i)}\}_{i=1}^{m} \sim \p_r$
        \State $\tilde \x \gets G_{\theta}(\x)$
        \For{$i \gets 1$ to $m$}
            \State Sample a random number $\epsilon \sim U(0, 1)$
            \State $\hat \x \gets \epsilon \x^{(i)} + (1 - \epsilon)\tilde \x^{(i)}$
            \State $\mathcal{L}_{\text{C}}^{(i)} \gets C_\psi(\x^{(i)}) - C_\psi(\tilde\x^{(i)}) + \lambda_{\text{C}} (\norm{\nabla_{\hat \x} C_\psi(\hat \x)}_2 - 1)^2$
        \EndFor
        \State $\psi \gets \text{Adam}(\nabla_\psi \frac{1}{m} \sum_{i=1}^{m} \mathcal{L}_{\text{C}}^{(i)}, \psi, \alpha, \beta_1, \beta_2)$
        \State $\mathcal{L}_{\text{G}} \gets \text{MSSIM}(\x, \tilde\x) + \lambda_{\text{G}}C_\psi(\tilde\x)$ 
        \State $\theta \gets \text{Adam}(\nabla_\theta \frac{1}{m} \sum_{i=1}^{m} -\mathcal{L}_{\text{G}}^{(i)}, \theta, \alpha, \beta_1, \beta_2)$
    \EndWhile
\end{algorithmic}}
\end{algorithm}

\begin{table*}[!t]
\setlength{\tabcolsep}{4pt}
\footnotesize	
\centering
\caption{ISCX-IDS dataset.}\label{tab:dataset_iscx}
    \begin{tabularx}{0.9\textwidth}{@{}TTTT@{}} 
        \toprule
        \multicolumn{2}{c}{\textbf{{Normal}}} & \multicolumn{2}{c}{\textbf{{Anomaly}}} \\
        \cmidrule(r){1-2} \cmidrule(r){3-4} 
         \textbf{{Traffic Type}} & \textbf{{\# of Flows}} & \textbf{{Traffic Type}} & \textbf{{\# of Flows}} \\
        \midrule 
         \multirow{4}{3cm}{HTTP, SMTP, SSH, IMAP, POP3, and FTP} & \multirow{4}{*}{869,978} & Infiltration & 9,925 \\
            &  & HTTP DoS & 3,427 \\   
            &  & DDoS & 21,129 \\     
            &  & Brute Force SSH & 6,964 \\  
        \bottomrule
    \end{tabularx}
\end{table*}  

\begin{table*}[!t]
\setlength{\tabcolsep}{4pt}
\footnotesize	
\centering
\caption{USTC-TFC dataset.}\label{tab:dataset_ustc}
    \begin{tabularx}{0.9\textwidth}{@{}TTTTT@{}} 
        \toprule
        \multicolumn{3}{c}{\textbf{{Normal}}} & \multicolumn{2}{c}{\textbf{{Anomaly}}} \\
        \cmidrule(r){1-3} \cmidrule(r){4-5} 
        \textbf{{App}} & \textbf{{Traffic Type}} & \textbf{{\# of Flows}} & \textbf{{Traffic Type}} & \textbf{{\# of Flows}} \\
        \midrule 
        Bittorrent & P2P & 15,00 & Cridex & 24,581 \\
        Facetime & Voice/Video & 6,000 & Geodo & 47,666 \\
        FTP & Data transfer & 202,034 & Htbot & 12,652 \\
        Gmail & Email/Webmail & 17,178 & Miuref & 20,755 \\
        MySQL & Database & 172,114 &  Neris & 44,605 \\
        Outlook & Email/Webmail & 14,984 & Nsis-ay & 11,014 \\
        Skype & Chat/IM & 12,000 & Shifu & 15,766 \\
        SMB & Data transfer & 77,781 & Tinba & 16,208 \\
        Weibo & Social Network & 79,810 & Virut & 58,638 \\
        World of Warcraft & Game & 15,761 & Zeus & 21,741 \\
        \bottomrule
    \end{tabularx}
\end{table*}  

\begin{table}[!t]
\setlength{\tabcolsep}{4pt}
\footnotesize	
\centering
\caption{MIRAI-RGU dataset.}\label{tab:dataset_rgu}
    \begin{tabularx}{0.9\linewidth}{TT} 
        \toprule
        \textbf{{Normal Traffic Type}} & \textbf{{\# of Flows}} \\                
        \midrule
        HTTP & 3,526,212 \\
        \midrule
        \textbf{{Anomaly Traffic Type}} & \textbf{{\# of Flows}} \\                
        \midrule
        Infection & 2,795,422 \\ 
        GREETH Flood & 67,116 \\ 
        VSE Flood & 4,990 \\ 
        ACK Flood & 137,838 \\
        DNS Flood & 9,704 \\ 
        HTTP Flood & 272 \\ 
        UDP PLAIN Flood & 18 \\ 
        UDP Flood  & 32,418 \\
        SYN Flood & 47,682 \\ 
        GREIP Flood & 77,293 \\
        \bottomrule
    \end{tabularx}
\end{table}  

\subsection{Adversarial Training}\label{sec:adv_training}
We address the generalization problem by regularizing the \ac{AE} through adversarial training. Additionally to maximizing MSSIM, we further maximize the reconstruction scores provided by the critic $C$. By doing so, besides generating contextually similar reconstructions, the \ac{AE} must reconstruct normal-like samples as faithfully as possible so the scores given by the critic $C$ are maximized. During training, the \ac{AE} is optimized to maximize 
\begin{equation}\label{eq:gen_ae}
    \mathcal{L}_{\text{G}} = \EX_{\x \sim \p_r} \big[ \text{MSSIM}(\x, \tilde\x) + \lambda_{\text{G}}C(\tilde\x)\big],
\end{equation}
where $\lambda_{\text{G}}$ is the regularization coefficient that balance the terms of the \ac{AE}'s objective function.

In Equation~\eqref{eq:gen_ae}, it is assumed that critic $C$ can provide high scores for real normal traffic samples and low scores for reconstruction. To do so, the critic $C$ must learn the normal and reconstruction data distributions. Therefore, during training, the critic $C$ is optimized to maximize 
\begin{equation}\label{eq:critic}
    \mathcal{L}_{\text{C}} = \EX_{\x \sim \p_r} \big[C(\x) - C(\tilde\x)\big] + \lambda_{\text{C}} \mathcal{L}_{\text{GP}},
\end{equation}
where $\mathcal{L}_{\text{GP}}$ is given by Equation~(\ref{eq:gp}), and $\lambda_{\text{C}} = 10$ as suggested in~\cite{gulrajani2017improved}. Our adversarial training strategy is based on the WGAN-GP framework described in Section~\ref{sec:wgan}. Algorithm~\ref{pseudo:algo1} summarizes the essential steps of the proposed adversarial training.  

\subsection{Anomaly Score}\label{sec:anom_score}
An anomaly score $\mathcal{A}(\x)$ is a function that provides a score to a sample $\x$ in the test set concerning samples in the training set. Samples with more significant anomaly scores are considered more likely to be anomalous. Traditionally, \ac{AE} strategies for anomaly detection rely on the reconstruction error between the input, and the reconstruction of the input has been used as an anomaly score. Another widely adopted anomaly score is the feature matching error based on an intermediate layer of the discriminator~\cite{schlegl2017unsupervised,truong2020empirical}. 

Experiments with the feature matching error as an anomaly score did not significantly improve ARCADE's performance. At the same time, it considerably increased the inference time since it is required to feed $\x$ and $\tilde\x$ through $C$ for feature extraction. Similarly, we found that using MSSIM as an anomaly score leads to a more discriminative anomaly function when compared to $\mathcal{L}_2$. However, the gains in efficiency are not meaningful enough to justify the loss in efficiency due to the SSIM's complexity. Therefore, for a given sample $\x$ in the test set, its anomaly score computed using ARCADE is denoted as $\mathcal{A}(\x) = \mathcal{L}_2(\x, \tilde \x)$.

\section{Experimental Evaluation}\label{sec:experiments}
The present section investigates and compares the performance of ARCADE with baselines on three network traffic datasets. The considered datasets and baselines are described in Section~\ref{sec:datasets} and Section~\ref{sec:baselines}, respectively. Implementation, training details, and hyper-parameter tuning are described in Section~\ref{sec:training_det}. In Section~\ref{sec:exp_effectiviness}, we assess the effectiveness of ARCADE and baselines on the three considered datasets. Section~\ref{sec:complexity} present the analysis of the complexity and the detection speed of ARCADE and D-PACK baseline.

\subsection{Datasets Description}\label{sec:datasets}
We used three datasets to evaluate the performance of the proposed approach with real-world normal and malicious network traffic: ISCX-IDS~\cite{shiravi2012toward}, USTC-TFC~\cite{wang2017malware}, and MIRAI-RGU~\cite{mcdermott2018botnet}. The choice of datasets is based on the requirement for raw network traffic. The selected datasets are among the most well-known datasets for intrusion detection, which provide raw network traffic (\texttt{.pcap}) in addition to hand-designed features (\texttt{.csv}). For example, the KDD'99 and NSL-KDD datasets provide only hand-designed extracted features, which limits their use in this work. It is worth noting that the number of flows presented in dataset Table~\ref{tab:dataset_iscx}, \ref{tab:dataset_ustc}, and \ref{tab:dataset_rgu} described below are the number of flows achieved after the preprocessing procedure proposed in Section~\ref{sec:preprocessing}.

The ISCX-IDS dataset~\cite{shiravi2012toward} is a realistic-like dataset originally proposed for the development of enhanced intrusion detection and anomaly-based approaches. The network traffic was collected for seven days. Packets collected on the first and sixth days are normal traffic. Normal and attack packets are collected on the second and third days. In the fourth, fifth, and seventh days, besides the normal traffic, HTTP DoS, DDoS using an IRC Botnet, and \ac{BF} SSH packets are collected, respectively. Table~\ref{tab:dataset_iscx} provides an overview of the ISCX-IDS dataset. The USTC-TFC dataset~\cite{wang2017malware} includes ten classes of normal traffic and ten classes of malware traffic from public websites, which were collected from a real network environment from 2011 to 2015. Table~\ref{tab:dataset_ustc} provides an overview of the USTC-TFC dataset. The MIRAI-RGU dataset includes normal traffic from \ac{IoT} \ac{IP} cameras and ten classes of malicious traffic from the Mirai botnet malware, such as HTTP flood, UDP flood, DNS flood, Mirai infection traffic, VSE flood, GREIP flood, GREETH flood, TCP ACK flood, TCP SYN flood, and UDPPLAIN flood. Table~\ref{tab:dataset_rgu} provides an overview of the MIRAI-RGU dataset. 

We split each dataset into training, validation, and test sets. The training set is composed only of normal samples. Normal and anomaly samples are used only for testing and validation. We balance the test set such that each subset of classes in the test set presents the same number of samples. Note that the normal traffic in the test set is not a subset of the training set. The validation set is composed of 5\% of the samples of each class from the test set, randomly selected and removed for validation purposes.

\begin{figure}[!t]
    \centering
    \includegraphics[width=0.9\linewidth]{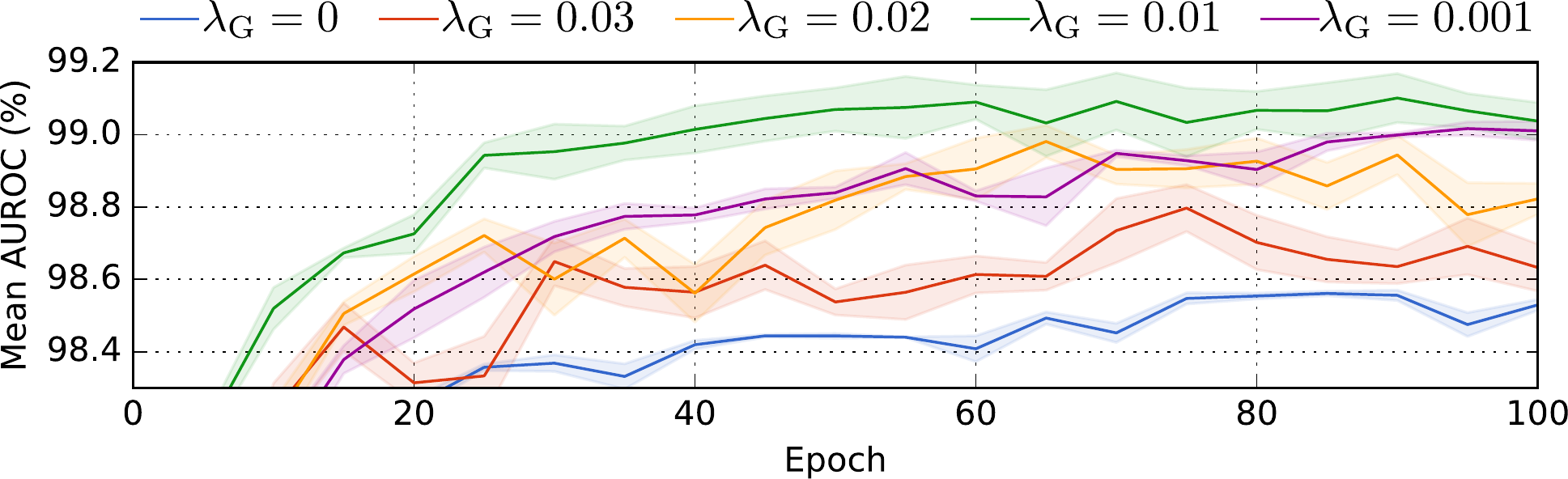}  
    \caption{ARCADE's mean AUROC (\%) convergence given varying values for the adversarial regularization coefficient $\lambda_{\text{G}}$. In the case where $\lambda_{\text{G}} = 0$, results are equivalent to the \ac{AE}-SSIM.}\label{fig:lambda2}
\end{figure} 

\begin{table}[!t]
\setlength{\tabcolsep}{4pt}
\footnotesize	
\centering
\caption{ARCADE's mean AUROC (\%) on the three considered datasets given varying input sizes. Results are in the format \textit{mean} $\pm$ \textit{std.} obtained over 10-folds.}\label{tab:exp_varying_n}
    \begin{tabularx}{\columnwidth}{@{}>{\hsize=2cm}XTTTTT@{}} 
        \toprule
        \multirow{2}{*}{\textbf{Dataset}} & \multicolumn{5}{c}{\textbf{Number of Packets}} \\
        \cmidrule{2-6} 
        & \textbf{2} & \textbf{3} & \textbf{4} & \textbf{5} & \textbf{6} \\
         
        \midrule 
        ISCX-IDS & 86.7$\pm$0.1 & 96.9$\pm$0.5 & 99.0$\pm$0.1 & 99.3$\pm$0.0 & 99.1$\pm$0.0\\
        USTC-TFC & 99.9$\pm$0.0 & 100\hphantom{.}$\pm$0.0 & 100\hphantom{.}$\pm$0.0 & 100\hphantom{.}$\pm$0.0 & 100\hphantom{.}$\pm$0.0 \\
        MIRAI-RGU & 99.6$\pm$0.0 &99.9$\pm$0.0 & 99.9$\pm$0.0 & 99.9$\pm$0.0 & 99.8$\pm$0.0\\
        \bottomrule
    \end{tabularx}
\end{table}

\begin{table}[!t]
\setlength{\tabcolsep}{4pt}
\footnotesize	
\centering

\caption{AUROC (\%) of ARCADE and shallow baselines. Results are in the format \textit{mean} $\pm$ \textit{std.} obtained over 10-folds. We present results for the ISCX-IDS with $n \in \{2,5\}$, denoted as ISCX-IDS$^n$.}\label{tab:exp_shallow}
    \begin{tabularx}{\columnwidth}{@{} >{\hsize=2cm}XTTTT @{}} 
        \toprule
        \multirow{2}{*}{\textbf{Dataset}} & \multicolumn{3}{c}{\textbf{Shallow Baselines}} & \textbf{Proposed} \\
        \cmidrule{2-5} 
        & \textbf{OC-SVM} & \textbf{KDE} & \textbf{IF} & \textbf{ARCADE} \\
         
        \midrule 
        ISCX-IDS$^2$ & 79.03$\pm$0.0 & 80.46$\pm$0.0 & 74.32$\pm$0.1 & \textbf{86.73}$\pm$0.1 \\
        ISCX-IDS$^5$ & 87.24$\pm$0.0 & 68.01$\pm$0.0 & 78.44$\pm$0.1 & \textbf{99.32}$\pm$0.0 \\
        USTC-TFC & 96.86$\pm$0.0 & 89.91$\pm$0.0 & 81.45$\pm$0.1 & \textbf{99.99}$\pm$0.0 \\
        MIRAI-RGU & 98.02$\pm$0.0 & 98.55$\pm$0.0 & 95.15$\pm$0.9 & \textbf{99.99}$\pm$0.0 \\
        \bottomrule
    \end{tabularx}
\end{table}

\subsection{Competing methods}\label{sec:baselines}
We compare ARCADE to three shallow and four deep learning methods for anomaly detection. The chosen shallow baselines are well-known methods typically applied to anomaly detection problems and commonly used as a benchmark for novel anomaly detection methods. D-PACK is the state-of-the-art unsupervised network anomaly detection method using raw network traffic bytes as input. We implemented and used D-PACK's effectiveness and efficiency as a baseline for ARCADE. GANomaly was chosen due to the similarities in the adversarial regularization strategies. Comparing ARCADE with GANomaly allows us to assess the effectiveness of our proposed adversarial strategy.
Moreover, comparing ARCADE with AE-$\ell_2$ and AE-SSIM allows us to confirm the findings presented by \citet{bergmann2018improving} and also assess the effectiveness of the proposed adversarial regularization since ARCADE with $\lambda_{\text{G}} = 0$ is equivalent to AE-SSIM. We also implemented and performed experiments with probabilistic models such as \ac{VAE} and \ac{AAE}; however, they did not produce satisfactory results when compared to deterministic \acp{AE}. Therefore, their results are not reported. Below we describe each competing method and its respective parameters.

\subsubsection{Shallow Baselines} (i)~One-Class SVM (OC-SVM)~\cite{scholkopf2001estimating} with Gaussian kernel. We optimize the hyperparameters $\gamma$ and $\nu$ via grid search using the validation set with $\gamma \in \{2^{-10}, 2^{-9}, \dots, 2^{0}\}$, and $\nu \in \{0.01, 0.02, \dots,  0.1\}$. (ii) Kernel density estimation (KDE). We optimize the bandwidth $h$ of the Gaussian kernel via grid search, given ten values spaced evenly between -1 to 1 on a logarithmic scale. (iii) Isolation Forest (IF)~\cite{liu2008isolation}. As recommended in the original work, we set the number of trees to 100 and the subsampling size to 256. For all three shallow baselines, we reduce the dimensionality of the data via \ac{PCA}, where we choose the minimum number of eigenvectors such that at least 95\% of the variance is retained.

\begin{table*}[!t]
\setlength{\tabcolsep}{4pt}
\footnotesize	
\centering
\caption{AUROC and F1-score (\%) of ARCADE and deep baselines. Each method was trained exclusively on normal network traffic, and the results are in the format mean ($\pm$ std.) obtained over 10-folds. For the ISCX-IDS, we run two experiments with $n\in\{2, 5\}$.}\label{tab:exp_datasets}
    \begin{tabularx}{\textwidth}{@{}>{\hsize=2.1cm}XT>{\hsize=0.75cm\hspace{-0.5cm}}TT>{\hsize=0.75cm\hspace{-0.5cm}}TT>{\hsize=0.75cm\hspace{-0.5cm}}TT>{\hsize=0.75cm\hspace{-0.5cm}}TT>{\hsize=0.75cm\hspace{-0.5cm}}T@{}} 
        \toprule
        \multirow{2.5}{2.1cm}{\textbf{ISCX-IDS}} & \multicolumn{2}{c}{\textbf{D-PACK}} & \multicolumn{2}{c}{\textbf{AE-$\boldsymbol{\ell_2}$}} & \multicolumn{2}{c}{\textbf{AE-SSIM}} & \multicolumn{2}{c}{\textbf{GANomaly}} & \multicolumn{2}{c}{\textbf{ARCADE}} \\
        \cmidrule{2-11}
        & \textbf{AUROC} & \textbf{F1} & \textbf{AUROC} & \textbf{F1} & \textbf{AUROC} & \textbf{F1} & \textbf{AUROC} & \textbf{F1} & \textbf{AUROC} & \textbf{F1} \\
        \midrule
        & \multicolumn{10}{c}{\textbf{Input Size 200}} \\
        \midrule
Infiltration & 99.36 ($\pm$0.1) & 99.07 & 99.20 ($\pm$0.1) & 99.04 & 99.13 ($\pm$0.0) & 99.10 & 99.12 ($\pm$0.0) & 98.98 & \textbf{99.43} ($\pm$0.0) & \textbf{99.16} \\
HTTP DoS & \textbf{83.63} ($\pm$2.8) & 67.53 & 77.59 ($\pm$0.3) & 61.96 & 77.98 ($\pm$0.4) & 62.15 & 79.63 ($\pm$0.5) & 66.06 & 81.22 ($\pm$2.2) & \textbf{68.70} \\
DDoS & 47.06 ($\pm$5.2) & 44.53 & 43.31 ($\pm$0.5) & 46.51 & 44.05 ($\pm$1.4) & 44.51 & 40.97 ($\pm$1.2) & 21.98 & \textbf{55.35} ($\pm$1.0) & \textbf{66.61} \\
BF SSH & 97.35 ($\pm$1.0) & 86.51 & 99.66 ($\pm$0.0) & 98.73 & 99.68 ($\pm$0.0) & 98.80 & 99.29 ($\pm$0.1) & 98.20 & \textbf{99.76} ($\pm$0.1) & \textbf{99.24} \\
\cmidrule{1-11}
All anomalies & 80.86 ($\pm$1.8) & 72.31 & 83.05 ($\pm$0.4) & 75.19 & 83.63 ($\pm$0.3) & 75.35 & 82.66 ($\pm$0.3) & 75.98 & \textbf{86.73} ($\pm$0.1) & \textbf{77.19} \\
        \midrule
        & \multicolumn{10}{c}{\textbf{Input Size 500}} \\
        \midrule
Infiltration & 96.68 ($\pm$2.4) & 93.88 & 99.32 ($\pm$0.0) & 99.01 & 99.35 ($\pm$0.0) & 99.09 & 99.26 ($\pm$0.0) & 98.69 & \textbf{99.62} ($\pm$0.0) & \textbf{99.17} \\
HTTP DoS & 93.56 ($\pm$2.0) & 88.99 & 92.48 ($\pm$0.0) & 90.84 & 92.28 ($\pm$0.1) & 90.67 & 92.59 ($\pm$0.3) & 91.56 & \textbf{93.72} ($\pm$0.4) & \textbf{91.95} \\
DDoS & \textbf{95.40} ($\pm$1.2) & \textbf{94.73} & 89.89 ($\pm$0.2) & 92.78 & 90.26 ($\pm$0.2) & 92.59 & 89.55 ($\pm$0.6) & 92.46 & 91.04 ($\pm$0.1) & 93.19 \\
BF SSH & 99.11 ($\pm$0.3) & 96.91 & 99.81 ($\pm$0.0) & 99.47 & 99.91 ($\pm$0.0) & 99.49 & \textbf{99.99} ($\pm$0.0) & \textbf{99.66} & 99.96 ($\pm$0.0) & 99.63 \\
        \cmidrule{1-11}
All anomalies & 96.38 ($\pm$2.5) & 93.14 & 98.58 ($\pm$0.0) & 96.72 & 98.62 ($\pm$0.0) & 96.90 & 98.58 ($\pm$0.0) & 97.22 & \textbf{99.32} ($\pm$0.0) & \textbf{97.29} \\
\midrule
        \multirow{2.5}{2.1cm}{\textbf{USTC-TFC}} & \multicolumn{2}{c}{\textbf{D-PACK}} & \multicolumn{2}{c}{\textbf{AE-$\boldsymbol{\ell_2}$}} & \multicolumn{2}{c}{\textbf{AE-SSIM}} & \multicolumn{2}{c}{\textbf{GANomaly}} & \multicolumn{2}{c}{\textbf{ARCADE}} \\
        \cmidrule{2-11}
        & \textbf{AUROC} & \textbf{F1} & \textbf{AUROC} & \textbf{F1} & \textbf{AUROC} & \textbf{F1} & \textbf{AUROC} & \textbf{F1} & \textbf{AUROC} & \textbf{F1} \\
\midrule
Cridex & 99.29 ($\pm$0.1) & 94.05 & \textbf{\hphantom{.}100\hphantom{0}} ($\pm$0.0) & \textbf{\hphantom{.}100\hphantom{0}} & \textbf{\hphantom{.}100\hphantom{0}} ($\pm$0.0) & \textbf{\hphantom{.}100\hphantom{0}} & 99.91 ($\pm$0.1) & 98.88 & \textbf{\hphantom{.}100\hphantom{0}} ($\pm$0.0) & \textbf{\hphantom{.}100\hphantom{0}} \\
Geodo & 99.28 ($\pm$0.2) & 94.08 & 99.99 ($\pm$0.0) & 99.91 & 99.99 ($\pm$0.0) & 99.88 & 99.77 ($\pm$0.0) & 96.35 & \textbf{\hphantom{.}100\hphantom{0}} ($\pm$0.0) & \textbf{\hphantom{.}100\hphantom{0}} \\
Htbot & 99.47 ($\pm$0.0) & 94.41 & \textbf{\hphantom{.}100\hphantom{0}} ($\pm$0.0) & \textbf{\hphantom{.}100\hphantom{0}} & \textbf{\hphantom{.}100\hphantom{0}} ($\pm$0.0) & \textbf{\hphantom{.}100\hphantom{0}} & 99.97 ($\pm$0.0) & 98.55 & \textbf{\hphantom{.}100\hphantom{0}} ($\pm$0.0) & \textbf{\hphantom{.}100\hphantom{0}} \\
Miuref & 99.42 ($\pm$0.1) & 94.63 & \textbf{\hphantom{.}100\hphantom{0}} ($\pm$0.0) & 99.97 & 99.99 ($\pm$0.0) & 99.86 & 99.83 ($\pm$0.0) & 97.73 & \textbf{\hphantom{.}100\hphantom{0}} ($\pm$0.0) & \textbf{\hphantom{.}100\hphantom{0}} \\
Neris & 99.76 ($\pm$0.0) & 95.05 & 99.99 ($\pm$0.0) & 99.83 & \textbf{\hphantom{.}100\hphantom{0}} ($\pm$0.0) & 99.97 & 99.98 ($\pm$0.0) & 99.37 & \textbf{\hphantom{.}100\hphantom{0}} ($\pm$0.0) & \textbf{\hphantom{.}100\hphantom{0}} \\
Nsis-ay & 99.72 ($\pm$0.1) & 93.89 & 99.99 ($\pm$0.0) & 99.94 & \textbf{\hphantom{.}100\hphantom{0}} ($\pm$0.0) & \textbf{\hphantom{.}100\hphantom{0}} & 99.99 ($\pm$0.0) & 99.41 & \textbf{\hphantom{.}100\hphantom{0}} ($\pm$0.0) & \textbf{\hphantom{.}100\hphantom{0}} \\
Shifu & 99.51 ($\pm$0.1) & 95.63 & \textbf{\hphantom{.}100\hphantom{0}} ($\pm$0.0) & \textbf{\hphantom{.}100\hphantom{0}} & \textbf{\hphantom{.}100\hphantom{0}} ($\pm$0.0) & \textbf{\hphantom{.}100\hphantom{0}} & 99.96 ($\pm$0.0) & 98.68 & \textbf{\hphantom{.}100\hphantom{0}} ($\pm$0.0) & \textbf{\hphantom{.}100\hphantom{0}} \\
Tinba & 99.92 ($\pm$0.0) & 96.12 & 99.99 ($\pm$0.0) & 99.97 & \textbf{\hphantom{.}100\hphantom{0}} ($\pm$0.0) & \textbf{\hphantom{.}100\hphantom{0}} & 99.99 ($\pm$0.0) & 99.91 & \textbf{\hphantom{.}100\hphantom{0}} ($\pm$0.0) & \textbf{\hphantom{.}100\hphantom{0}} \\
Virut & 99.80 ($\pm$0.1) & 95.96 & 99.99 ($\pm$0.0) & 99.91 & \textbf{\hphantom{.}100\hphantom{0}} ($\pm$0.0) & \textbf{\hphantom{.}100\hphantom{0}} & 99.99 ($\pm$0.0) & 99.45 & \textbf{\hphantom{.}100\hphantom{0}} ($\pm$0.0) & \textbf{\hphantom{.}100\hphantom{0}} \\
Zeus & 99.01 ($\pm$0.2) & 88.52 & \textbf{\hphantom{.}100\hphantom{0}} ($\pm$0.0) & \textbf{\hphantom{.}100\hphantom{0}} & \textbf{\hphantom{.}100\hphantom{0}} ($\pm$0.0) & \textbf{\hphantom{.}100\hphantom{0}} & 99.90 ($\pm$0.0) & 98.16 & \textbf{\hphantom{.}100\hphantom{0}} ($\pm$0.0) & \textbf{\hphantom{.}100\hphantom{0}} \\
        \cmidrule{1-11}
All anomalies & 99.59 ($\pm$0.2) & 98.77 & 99.99 ($\pm$0.0) & 99.89 & 99.99 ($\pm$0.0) & 99.93 & 99.81 ($\pm$0.0) & 99.40 & \textbf{99.99} ($\pm$0.0) & \textbf{99.98} \\
        \midrule
        \multirow{2.5}{2.1cm}{\textbf{MIRAI-RGU}} & \multicolumn{2}{c}{\textbf{D-PACK}} & \multicolumn{2}{c}{\textbf{AE-$\boldsymbol{\ell_2}$}} & \multicolumn{2}{c}{\textbf{AE-SSIM}} & \multicolumn{2}{c}{\textbf{GANomaly}} & \multicolumn{2}{c}{\textbf{ARCADE}} \\
        \cmidrule{2-11}
        & \textbf{AUROC} & \textbf{F1} & \textbf{AUROC} & \textbf{F1} & \textbf{AUROC} & \textbf{F1} & \textbf{AUROC} & \textbf{F1} & \textbf{AUROC} & \textbf{F1} \\
        \midrule
Infection & 99.66 ($\pm$0.1) & 98.33 & 99.74 ($\pm$0.0) & 99.68 & 99.77 ($\pm$0.0) & 99.83 & 99.80 ($\pm$0.0) & 99.47 & \textbf{99.99} ($\pm$0.0) & \textbf{99.83} \\
GREETH Flood & 99.77 ($\pm$0.1) & 99.47 & 99.96 ($\pm$0.0) & 99.85 & \textbf{99.98} ($\pm$0.0) & 99.88 & 99.97 ($\pm$0.0) & \textbf{99.88} & 99.97 ($\pm$0.0) & 99.86 \\
VSE Flood & 99.70 ($\pm$0.2) & 99.38 & 99.99 ($\pm$0.0) & 99.85 & 99.99 ($\pm$0.0) & 99.85 & 99.96 ($\pm$0.0) & 99.80 & \textbf{99.99} ($\pm$0.0) & \textbf{99.86} \\
ACK Flood & 99.90 ($\pm$0.0) & 99.41 & 99.99 ($\pm$0.0) & 99.81 & 99.99 ($\pm$0.0) & 99.89 & 99.99 ($\pm$0.0) & \textbf{99.89} & \textbf{\hphantom{.}100\hphantom{0}} ($\pm$0.0) & 99.89 \\
DNS Flood & 99.82 ($\pm$0.1) & 99.48 & 99.99 ($\pm$0.0) & 99.78 & 99.99 ($\pm$0.0) & 99.85 & 99.98 ($\pm$0.0) & 99.86 & \textbf{99.99} ($\pm$0.0) & \textbf{99.92} \\
HTTP Flood & \textbf{\hphantom{.}100\hphantom{0}} ($\pm$0.0) & 99.73 & \textbf{\hphantom{.}100\hphantom{0}} ($\pm$0.0) & 99.80 & \textbf{\hphantom{.}100\hphantom{0}} ($\pm$0.0) & 99.80 & \textbf{\hphantom{.}100\hphantom{0}} ($\pm$0.0) & 99.80 & \textbf{\hphantom{.}100\hphantom{0}} ($\pm$0.0) & \textbf{99.80} \\
UDP Plain Flood & \textbf{\hphantom{.}100\hphantom{0}} ($\pm$0.0) & \textbf{97.29} & \textbf{\hphantom{.}100\hphantom{0}} ($\pm$0.0) & \textbf{97.29} & \textbf{\hphantom{.}100\hphantom{0}} ($\pm$0.0) & \textbf{97.29} & \textbf{\hphantom{.}100\hphantom{0}} ($\pm$0.0) & \textbf{97.29} & \textbf{\hphantom{.}100\hphantom{0}} ($\pm$0.0) & \textbf{97.29} \\
UDP Flood & 99.69 ($\pm$0.2) & 99.33 & 99.96 ($\pm$0.0) & \textbf{99.86} & 99.94 ($\pm$0.0) & 99.83 & \textbf{99.99} ($\pm$0.0) & 99.85 & 99.96 ($\pm$0.0) & 99.84 \\
SYN Flood & 99.90 ($\pm$0.0) & 99.69 & 99.99 ($\pm$0.0) & 99.81 & \textbf{\hphantom{.}100\hphantom{0}} ($\pm$0.0) & 99.94 & 99.99 ($\pm$0.0) & 99.88 & \textbf{\hphantom{.}100\hphantom{0}} ($\pm$0.0) & \textbf{99.96} \\
GREIP Flood & 99.77 ($\pm$0.1) & 99.57 & \textbf{\hphantom{.}100\hphantom{0}} ($\pm$0.0) & \textbf{99.96} & \textbf{\hphantom{.}100\hphantom{0}} ($\pm$0.0) & 99.96 & \textbf{\hphantom{.}100\hphantom{0}} ($\pm$0.0) & 99.96 & \textbf{\hphantom{.}100\hphantom{0}} ($\pm$0.0) & 99.96 \\
        \cmidrule{1-11}
All anomalies & 99.72 ($\pm$0.1) & 99.81 & 99.92 ($\pm$0.0) & 99.96 & 99.93 ($\pm$0.0) & 99.97 & 99.92 ($\pm$0.0) & 99.94 & \textbf{99.99} ($\pm$0.0) & \textbf{99.97} \\
        \bottomrule
    \end{tabularx}
\end{table*}    

\subsubsection{Deep Baselines}\label{sec:deep_baselines}
(i)~D-PACK~\cite{hwang2020unsupervised}, recently proposed for unsupervised network anomaly detection, can be considered the state-of-the-art DL method for the task. D-PACK's performance serves as a point of comparison for ARCADE's effectiveness and efficiency. The original D-PACK formulation assumes that normal traffic is split into multiple classes, as in the USTC-TFC dataset. However, this is not the circumstance for most public datasets, such as the other two datasets considered here. We empirically assessed that removing the softmax classifier degrades the method's efficiency. Therefore, we keep the original D-PACK formulation even for datasets without labeled normal training data. The network architecture, training strategy, and hyperparameters were kept as recommended in the original work. 
(ii)~GANomaly~\cite{akcay2018ganomaly} was originally proposed for image anomaly detection. Here, we do not employ it as an out-of-the-box anomaly detection approach. However, we use its adversary training framework with the proposed 1D-CNN model architecture presented in Section~\ref{sec:architecture}. The idea is to compare GANomaly's training strategy with our proposed adversarial training strategy. Note that GANomaly defines the generator $G$ as an encoder-decoder-encoder. Therefore, a second encoder $E'$ with the same architecture of $E$ (without sharing parameters) is added to the proposed \ac{AE}, where the input of $E'$ is the outcome of the decoder $D$, i.e., the input for encoder $E'$ is the reconstruction of the input. Finally, we modify the critic $C$ to align with their proposed discriminator $D$. We modify $C$, so batch normalization is used instead of layer normalization, and a Sigmoid activation function is added after the last layer. The anomaly score is given by the $\ell_2$-distance between the latent space of $E$ and the latent space of $E'$. We performed grid search optimize $w_{\text{rec}} \in \{50, 75, 100, 125, 150\}$ and results suggest that $w_{\text{rec}} = 75$ lead to best results. All the other parameters were kept as suggested in the original work. 
(iii)~\ac{AE}-$\ell_2$ is an \ac{AE} with the same proposed network architecture in Section~\ref{sec:architecture}, where $\mathcal{L}_2$ loss is used as distance metric during training, and $\mathcal{L}_2$ is also used for the anomaly score computation.
(iv)~\ac{AE}-SSIM is an \ac{AE} with the same proposed network architecture in Section~\ref{sec:architecture}, where MSSIM loss is used for training, and $\mathcal{L}_2$ is used for computing the anomaly scores. In this work, we used the PyTorch Image Quality (PIQ)~\cite{piq} implementation of the SSIM loss with Gaussian kernel and kernel size $K=3$, obtained through a grid search optimization with $K \in \{3, 5, 7, 9\}$. 
  
\begin{table*}[!t]
\setlength{\tabcolsep}{4pt}
\footnotesize	
\centering
\caption{The accuracy, precision, recall, and F1-score values in \% of ARCADE and D-PACK for the 99th percentile and maximum thresholds. Results are in the format of the mean ($\pm$ std.) obtained over ten seeds.}\label{tab:exp_threshold}
    \begin{tabularx}{\textwidth}{@{}>{\hsize=2.0cm}XTTTT>{\hsize=0.1cm}TTTTT@{}} 
        \toprule
        \multirow{2}{*}{\textbf{Dataset}} & \multicolumn{4}{c}{\textbf{D-PACK}} & & \multicolumn{4}{c}{\textbf{ARCADE}} \\
        \cmidrule{2-5}  
        \cmidrule{7-10}  
         & \textbf{Accuracy} & \textbf{Precision} & \textbf{Recall} & \textbf{F1-score} & & \textbf{Accuracy} & \textbf{Precision} & \textbf{Recall} & \textbf{F1-score} \\
        \midrule  
        & \multicolumn{8}{c}{\textbf{99th percentile threshold}} \\
        \midrule     
ISCX-IDS & 75.91 ($\pm$0.08) & 97.98 ($\pm$0.01) & 52.83 ($\pm$0.15) & 67.43 ($\pm$0.13) & & \textbf{93.07} ($\pm$0.01) & \textbf{98.86} ($\pm$0.00) & \textbf{87.15} ($\pm$0.01) & \textbf{92.63} ($\pm$0.01) \\
USTC-TFC & 97.71 ($\pm$0.00) & 98.97 ($\pm$0.00) & 96.43 ($\pm$0.01) & 97.68 ($\pm$0.00) & & \textbf{99.49} ($\pm$0.00) & \textbf{99.00} ($\pm$0.00) & \textbf{\hphantom{.}100\hphantom{0}} ($\pm$0.00) & \textbf{99.50} ($\pm$0.00) \\
Mirai-RGU & 98.50 ($\pm$0.01) & 99.87 ($\pm$0.00) & 98.44 ($\pm$0.02) & 99.14 ($\pm$0.01) & & \textbf{99.89} ($\pm$0.00) & \textbf{99.87} ($\pm$0.00) & \textbf{\hphantom{.}100\hphantom{0}} ($\pm$0.00) & \textbf{99.93} ($\pm$0.00) \\
    \midrule
    Mean & 90.70 & 98.94 & 82.56 & 87.08 & & \textbf{97.48} & \textbf{99.24} & \textbf{95.71} & \textbf{97.35} \\
    \midrule  
        & \multicolumn{8}{c}{\textbf{Maximum threshold}} \\
        \midrule     
ISCX-IDS & 50.00 ($\pm$0.00) & \textbf{\hphantom{.}100\hphantom{0}} ($\pm$0.00) & 0.016 ($\pm$0.00) & 0.033 ($\pm$0.00) & & \textbf{60.11} ($\pm$0.04) & \textbf{\hphantom{.}100\hphantom{0}} ($\pm$0.00) & \textbf{20.21} ($\pm$0.07) & \textbf{33.01} ($\pm$0.10) \\
USTC-TFC & 50.83 ($\pm$0.00) & \textbf{\hphantom{.}100\hphantom{0}} ($\pm$0.00) & 1.627 ($\pm$0.00) & 3.202 ($\pm$0.00) & & \textbf{99.88} ($\pm$0.00) & \textbf{\hphantom{.}100\hphantom{0}} ($\pm$0.00) & \textbf{99.77} ($\pm$0.00) & \textbf{99.88} ($\pm$0.00)  \\
Mirai-RGU & 32.67 ($\pm$0.11) & \textbf{\hphantom{.}100\hphantom{0}} ($\pm$0.00) & 24.45 ($\pm$0.12) & 37.75 ($\pm$0.16) & & \textbf{76.52} ($\pm$0.02) & \textbf{\hphantom{.}100\hphantom{0}} ($\pm$0.00) & \textbf{73.65} ($\pm$0.02) & \textbf{84.81} ($\pm$0.01)  \\
    \midrule
    Mean & 44.5 & \textbf{100} & 8.69 & 13.66 & & \textbf{78.83} & \textbf{100} & \textbf{64.54} & \textbf{72.56} \\    
    \bottomrule
    \end{tabularx}
\end{table*}

\begin{figure*}[!t]
    \centering
    \setlength\tabcolsep{1pt}
    \begin{tabular}{cc}
    \footnotesize{\textbf{D-PACK}} & \footnotesize{\textbf{ARCADE}} \\ [-1mm]
    \includegraphics[width=0.45\linewidth]{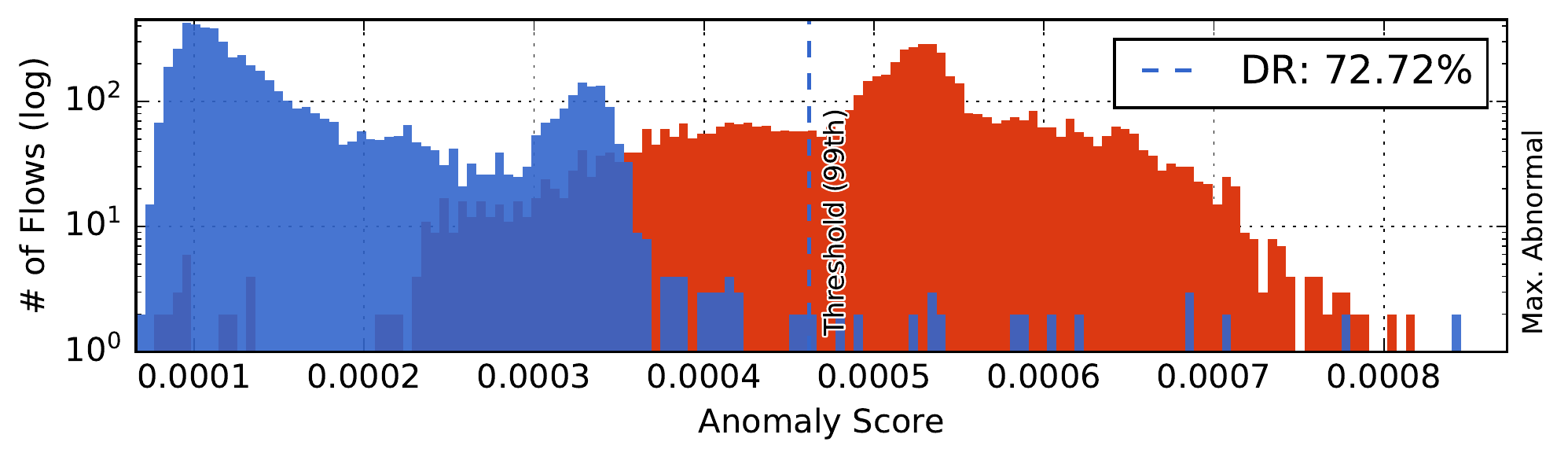} & \includegraphics[width=0.45\linewidth]{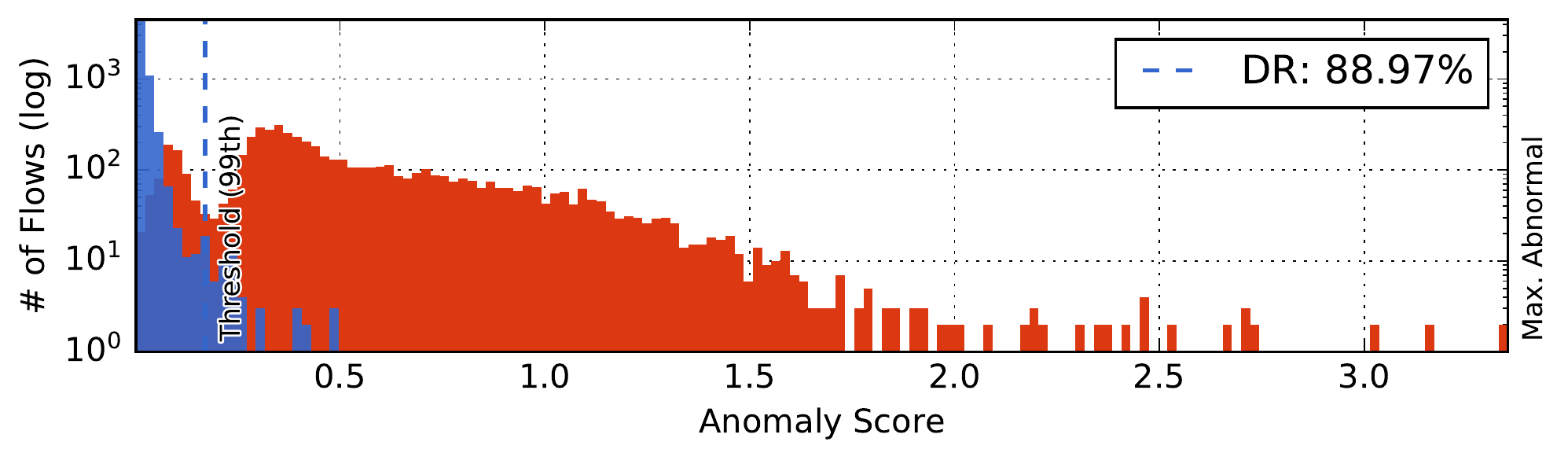}  \\ [-2mm]
    \multicolumn{2}{c}{\footnotesize{(a) ISCX-IDS}} \\ [1mm]
    \includegraphics[width=0.45\linewidth]{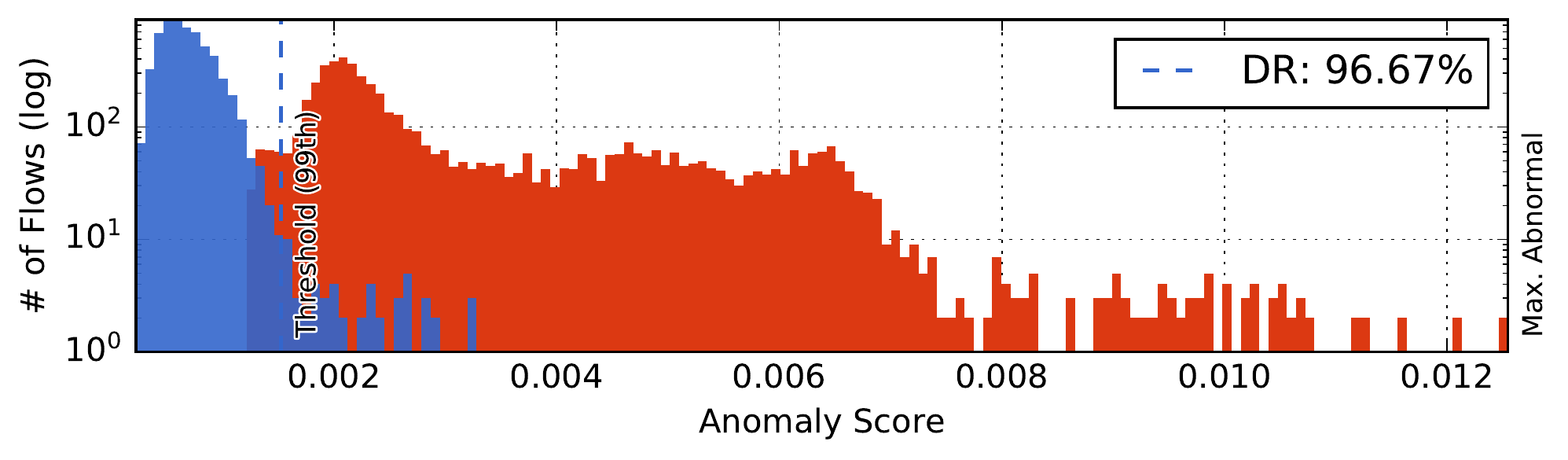} & \includegraphics[width=0.45\linewidth]{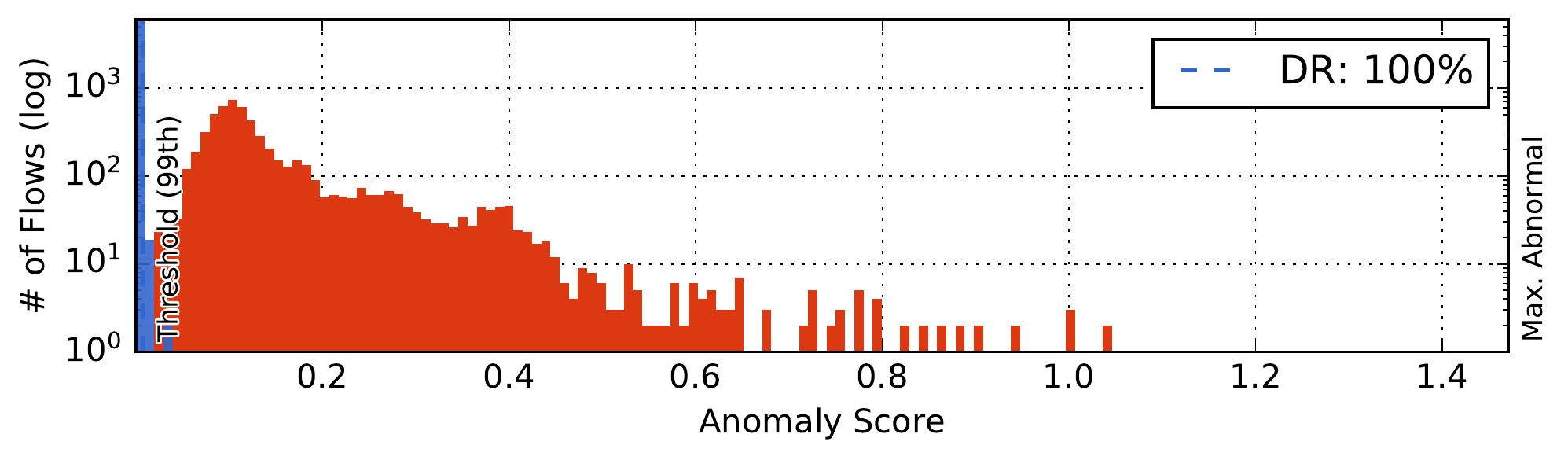}  \\ [-2mm]
    \multicolumn{2}{c}{\footnotesize{(b) USTC-TFC}} \\ [1mm]
    \includegraphics[width=0.45\linewidth]{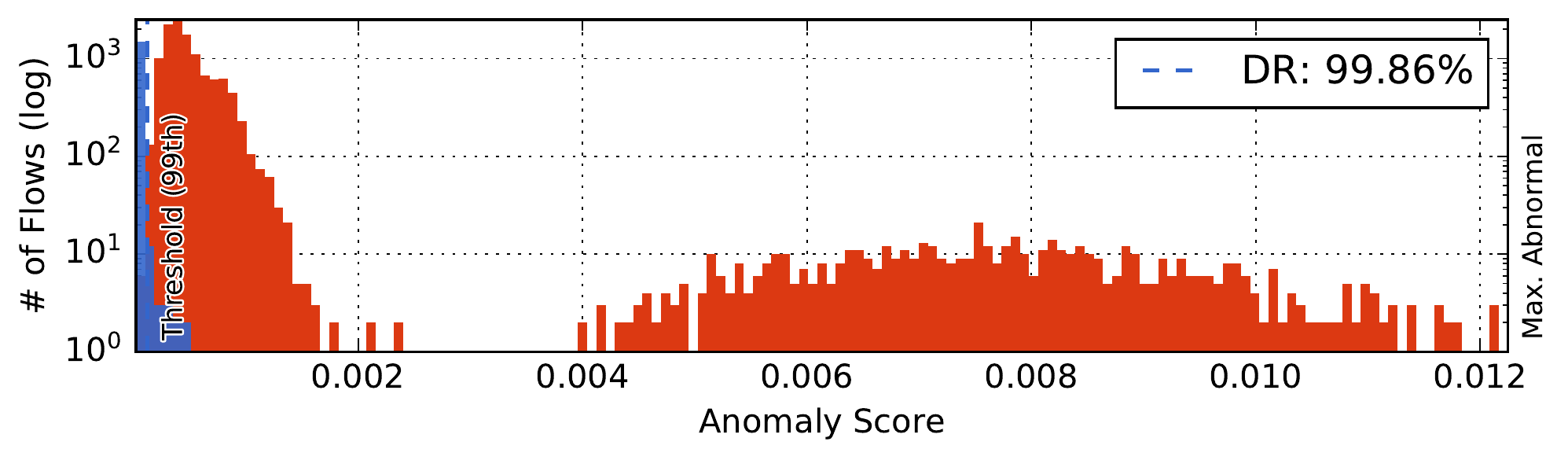} & \includegraphics[width=0.45\linewidth]{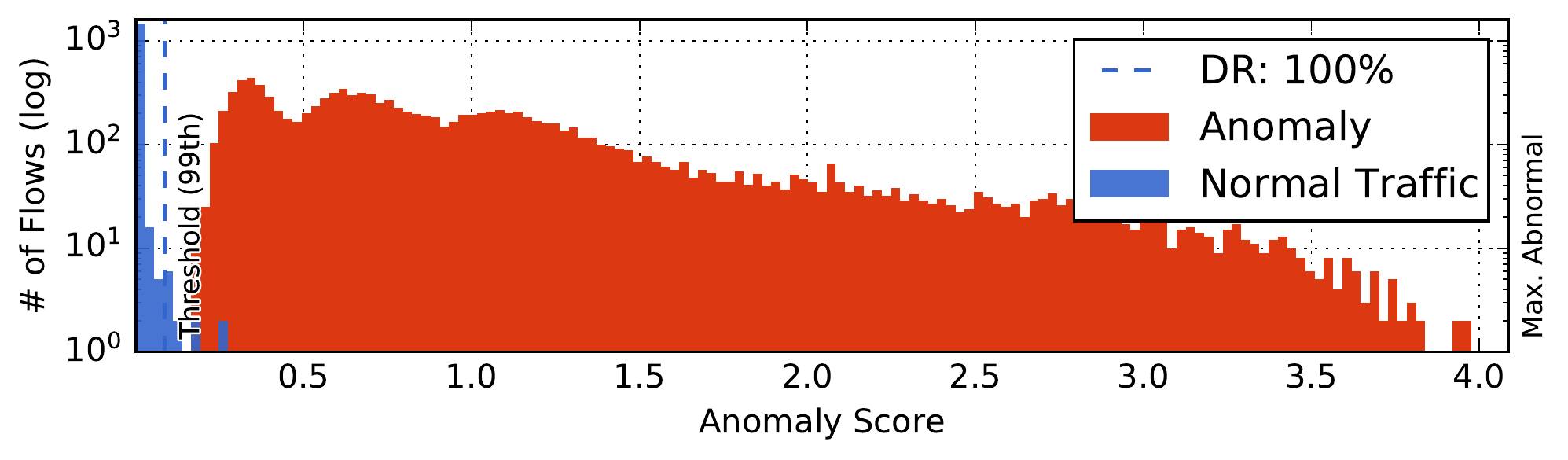}  \\[-2mm]
    \multicolumn{2}{c}{\footnotesize{(c) MIRAI-RGU}} \\
    \end{tabular} 
    \caption{The anomaly score distributions for normal and abnormal traffic from the test set of each considered dataset. Anomaly scores were computed using the best model’s parameters over 10-folds for each method. The DR was calculated based on the 99th percentile threshold of the normal traffic scores. Blue and red bars represent normal and abnormal traffic flows, respectively.}\label{fig:histograms}
\end{figure*}

\subsection{Training Recipe and Ablation Study}\label{sec:training_det}
The training objective (described in Section~\ref{sec:adv_training}) is optimized via Adam optimizer~\cite{kingma2014adam} with $\alpha=\text{1e-4}$, $\beta_1 = 0$, and $\beta_2=0.9$. It is worth noting again that Algorithm~\ref{pseudo:algo1} describes the main steps of the proposed adversarial training procedure. Additionally, we employ for all approaches a two-phase (``searching" and ``fine-tuning") learning rate $1\mathrm{e}{-4}$ for 100 epochs. In the fine-tuning phase, we train with the learning rate $1\mathrm{e}{-5}$ for another 50 epochs. The latent size $d$ is computed with PCA, equivalent to the minimum number of eigenvectors such that the sum of their explained variance is at least 95\%, i.e., $d \approx 50$ with $n=2$ for all three datasets. 

Given the hyperparameters and training recipe above, we performed ablation experiments to assess the performance of ARCADE with and without the proposed adversarial regularization and varying input sizes. The validation set was used for the ablation experiments. To assess the effectiveness of the proposed adversarial regularization, we performed experiments with the ICSX-IDS dataset with $n=5$, and $\lambda_{\text{G}} \in \{0, 0.001, 0.01, 0.02, 0.03\}$. Figure~\ref{fig:lambda2} illustrates the mean AUROC convergence (lines) and standard deviation (error bars amplified 50 times for visualization purposes). We can verify that the proposed adversarial-based regularization improves the capabilities of the \ac{AE} for network anomaly detection concerning the same \ac{AE} without the adversarial regularization, i.e., with $\lambda_{\text{G}} = 0$ ARCADE is equivalent to \ac{AE}-SSIM. The proposed adversarial training strategy can be exploited to improve the network anomaly detection capabilities of similar DL approaches, especially for scenarios where increasing the model size is not an option due to hardware constraints. Based on these results, we fix the adversarial regularization coefficient to $\lambda_{\text{G}} = 0.01$ for all the following experiments. We also analyze the ARCADE performance given different input sizes. Table~\ref{tab:exp_varying_n} presents the mean AUROC and standard deviations on the three datasets with $n \in \{2, 3, 4, 5, 6\}$. ARCADE achieves nearly 100 AUROC with $n=2$ on the USTC-TFC and MIRAI-RGU datasets. For the ISCX-IDS dataset, the method achieves 86.7 and 99.1 AUROC with 2 and 4 packets, respectively. The following experiments further investigate the considerable difference in performance given varying values of $n$. For the MIRAI-RGU dataset, the AUROC decreases with $n > 5$. Scaling the model depth and width given the input size could help since, for larger input sizes, more layers and channels would lead to an increased receptive field and more fine-grained patterns. Note that additional ablation experiments concerning the $\ell_2$ and \ac{SSIM} loss functions are provided in the following section.
  
\subsection{Network Traffic Anomaly Detection Results}\label{sec:exp_effectiviness}
We now systematically compare the proposed ARCADE's effectiveness with the baselines. 
Table~\ref{tab:exp_shallow} presents the results of the considered shallow baselines on the three network traffic datasets. ARCADE outperforms all of its shallow competitors. Table~\ref{tab:exp_datasets} presents the results of ARCADE and considered deep baselines. Here, we expand the evaluations to include a one-class anomaly detection setting, where each anomaly class is evaluated separately. Therefore, the table also includes the AUROC and F1-score concerning the evaluation performed exclusively on each anomaly class presented in each dataset. Note that the anomaly samples used for this evaluation are not necessarily a subset of the test set and were fixed for all methods. This allows each method to be evaluated separately against each attack or malware in each dataset.

The results for the deep baselines, considering normal and all anomalies, show that ARCADE outperforms all other methods on the three considered datasets. 
The methods rank ARCADE, AE-SSIM, AE-$\ell_2$, GANomaly, and D-PACK for results on the ISCX-IDS with $n=2$, USTC-TFC with $n=2$, and MIRAI-RGU with $n=2$. In experiments with the ISCX-IDS with $n=5$, the methods rank ARCADE, GANomaly, AE-SSIM, AE-$\ell_2$, and D-PACK. Despite having approximately 20 times more parameters than the proposed model, D-PACK achieved the worst results among the deep baselines. 
Results for the AE-SSIM and AE-$\ell_2$, similarly to the results provided in~\cite{bergmann2018improving}, show that using SSIM as a distance metric during training can improve the \ac{AE}'s capabilities in detecting anomalies. ARCADE, which also uses SSIM as a distance metric during training and employs the proposed adversarial regularization strategy, achieved better results than AE-SSIM, emphasizing the advantages of the proposed adversarial training strategy.
The GANomaly framework, comprised of its distinct model architecture, adversarial training strategy, and anomaly score, did not achieve better results than ARCADE. It is worth noting that GANomaly used the same \ac{AE} architecture as ARCADE with the requirement of an additional encoder, as described in Section~\ref{sec:deep_baselines}. 

The isolated validations for the ISCX-IDS with $n=2$ show that ARCADE achieved the best F1-score values for all classes and best AUROC values for Infiltration, DDoS, and \ac{BF} SSH, where D-PACK achieved the best AUROC for HTTP DoS. With $n=5$, ARCADE achieved the best results for Infiltration and HTTP DoS, where D-PACK achieved the best results for DDoS, and GANomaly achieved the best results for \ac{BF} SSH. The low performance of the considered methods on DoS and DDoS with $n=2$ indicates that analyzing the spatial relation between bytes among multiple subsequent packets is essential to detect such attacks. A single packet of a flood attack does not characterize an anomaly; e.g., an SYN packet can be found within the normal traffic. However, multiple SYN packets in sequence can be characterized as an SYN flood attack. This indicates that the spatial relation between bytes among subsequent packets is essential to detect such attacks.
In isolated experiments with anomaly classes from the USTC-TFC dataset, ARCADE achieved maximum results with 100 AUROC and 100 F1-score in all malware classes. Results from the isolated experiments with anomaly classes from the MIRAI-RGU show that, if we consider D-PACK, AE-$\ell_2$, AE-SSIM, and GANomaly, there is no clear winner. ARCADE achieved the best AUROC and F1-score values on the 8 and 6 classes, respectively. GANomaly ranked second with four best AUROC and three best F1-score values.

In practice, a threshold value must be set to distinguish between normal and anomalous traffic based on the anomaly score distribution of the normal traffic. In a supervised scenario where the normal and known anomalies' anomaly score distribution does not overlap, the maximum anomaly score of the normal traffic can lead to 100\% \ac{DR} and 0\% \ac{FAR}. This is commonly adopted since it leads to small \ac{FAR}. To avoid the impact of extreme maximum anomaly scores, the 99th percentile of the anomaly score distribution of the normal traffic can be used as an alternative. The downside of this approach is that approximately 1\% \ac{FAR} is expected. Regardless, the definition of the threshold is problem-dependent and is strongly related to \ac{IDS} architecture altogether, e.g., in a hybrid \ac{IDS} (anomaly-based and signature-based), where the anomaly-based method is used as a filter to avoid unnecessary signature verification, a high threshold could lead to low detection rates. In this case, a lower threshold, such as the 99th percentile (or even smaller), would be preferable since the signature-based approach would further validate false positives.  

We further compare ARCADE and D-PACK considering accuracy, precision, recall, and F1-score given two thresholds: (i)~the 99th percentile, and (ii)~the maximum value of the normal traffic anomaly scores. This comparison aims to analyze the effectiveness of ARCADE compared to the D-PACK baseline, originally proposed for network anomaly detection. The other deep baselines use the same model architecture as ARCADE and can be seen as contributions to this work we implemented.
Table~\ref{tab:exp_threshold} presents the accuracy, precision, recall, and F1-score of ARCADE and D-PACK given both thresholds, with $n=2$ for the USTC-TFC and MIRAI-RGU datasets, and $n=5$ for the ISCX-IDS dataset. The results of the 99th threshold show that ARCADE achieved the highest recall rate for the USTC-TFC and MIRAI-RGU datasets. This is because ARCADE produced no false negatives. ARCADE achieved an 11.79\% higher F1-score than D-PACK. When the maximum threshold is used, the ARCADE enhancement in performance is more clearly seen. As expected, both approaches were able to achieve the highest precision. However, D-PACK only achieved 8.69\% mean recall, while ARCADE achieved 64.54\%. This is an improvement of 642.69\%. 
Figure~\ref{fig:histograms} shows the anomaly score distribution of ARCADE and D-PACK computed using the model parameters that led to the best AUROC obtained over 10-folds on the three datasets. The detection rate is reported and calculated using the 99th percentile threshold, also presented in the figures. When considering the best model parameters and a 99th percentile threshold, ARCADE outperformed D-PACK in terms of detection rates by 22.35\%, 3.44\%, and 0.14\% on the ISCX-IDS, USTC-TFC, and Mirai-RGU datasets, respectively.

\subsection{Model Complexity and Detection Speed}\label{sec:complexity}
\begin{figure}[!t]
    \centering        
    \includegraphics[width=0.9\linewidth]{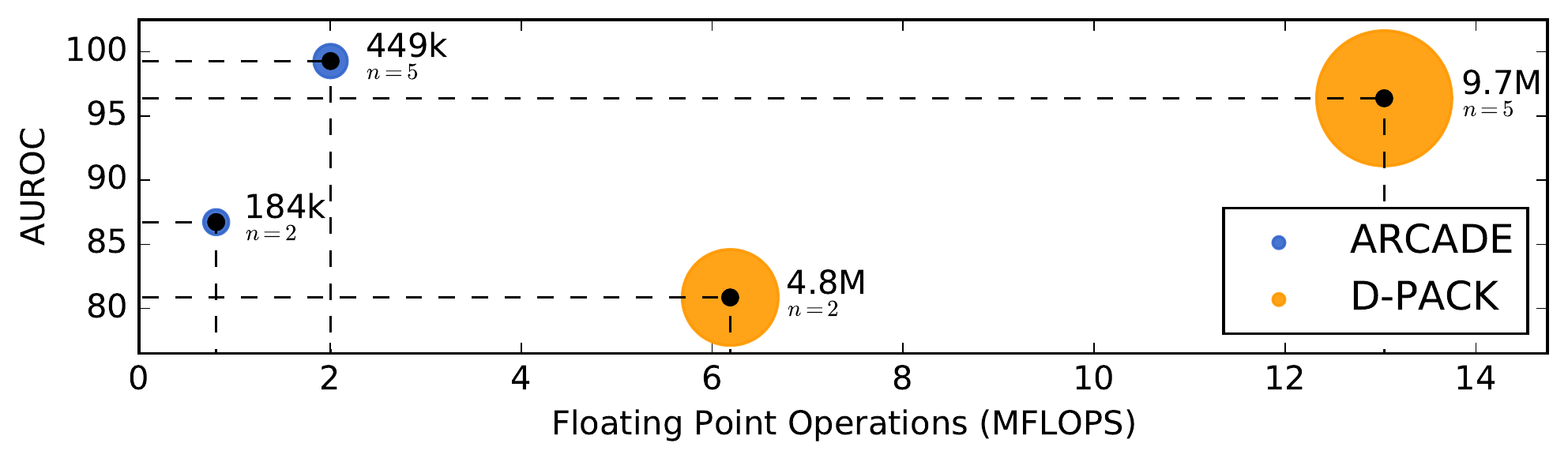} 
    \caption{Comparison between efficiency, effectiveness, and model size of ARCADE and D-PACK. We report AUROC (\%) vs.\ floating-point operations (FLOPS) required for a single forward pass is reported with $n\in\{2, 5\}$. The size of each circle corresponds to the model size (number of parameters). ARCADE achieves higher AUROC with approximately 20 times fewer parameters than D-PACK.}
    \label{fig:efficiencygflop}
\end{figure}

Here we evaluate the efficiency of ARCADE by comparing with D-PACK the number of samples processed per second, model sizes, and floating-point operations (FLOPS). Figure~\ref{fig:efficiencygflop} presents ARCADE and D-PACK efficiency, effectiveness, and model size on the ISCX-IDS with $n=2$ and $n=5$, where our ARCADE significantly outperforms D-PACK in all evaluated measures. We analyze the detection speed performance of ARCADE and D-PACK by assessing how many samples per second they can process in different environments with distinct processing capabilities that we categorize as edge, fog, and cloud. The device specifications and the experimental environment are summarized in Table~\ref{tab:spec}. We consider a Raspberry Pi 4B as an edge device, UP Xtreme and Jetson Xavier NX as fog devices, and a desktop personal computer with an AMD Ryzen Threadripper 3970X 32-core CPU, NVIDIA GeForce RTX 3090 GPU, and 128 GB RAM as a cloud device. Detection speed experiments were conducted with and without GPU support to account for the fact that edge (and sometimes fog) nodes may not have a GPU device, as is the case with the Raspberry Pi 4 and the UP Xtreme board. The NVIDIA Jetson Xavier NX and the personal computer were given a GPU warm-up stage of 5 seconds immediately before starting the experiment. The mean amount of processed flows per second was computed given ten runs. All experiments were implemented in Python 3.8 PyTorch version 1.8 without any improvement to speed up inference. 
Table~\ref{tab:det_speed} present the detection speed results with $n=2$. 
The results show that ARCADE outperformed D-PACK in all environments, with ARCADE being approximately 8, 3, 2.8, 2, 2.16 times faster on the Raspberry Pi 4, UP Xtreme, NVIDIA Jetson, Threadripper CPU, and RTX 3090 GPU, respectively. ARCADE can process over 1.9M flows per second on the RTX 3090 GPU. 
The definition of an ``optimal model" in an online network detection scenario cannot be well-defined since there is a clear trade-off between the model's effectiveness and complexity. In this sense, the proposed model can be easily adapted by changing the number of layers and channels, together with the input size, to better suit the needs of a particular environment given its processing power capabilities.

\begin{table}[!t]
\parnoteclear
\setlength{\tabcolsep}{4pt}\centering\footnotesize	
\caption{Mean detection speed comparison between ARCADE and D-PACK.}\label{tab:det_speed}
    \begin{tabularx}{\columnwidth}{@{}>{\hsize=3.75cm}XTT@{}} 
        \toprule
        \multirow{2}{*}{\textbf{Device}} & \multicolumn{2}{c}{\textbf{Detection Speed (flows/s)}} \\
        \cmidrule{2-3} 
        & \textbf{D-PACK} & \textbf{ARCADE} \\ 
        \midrule 
        Raspberry Pi 4 Model B          & 79        & 661       \\
        UP Xtreme WHLI7-A20-16064       & 4,120     & 12,471    \\
        NVIDIA Jetson Xavier NX         & 23,270    & 66,737    \\
        Ryzen Threadripper 3970X    & 22,659    & 47,478    \\
        NVIDIA GeForce RTX 3090         & 888,492   & 1,926,169 \\
        \bottomrule
    \end{tabularx}
    \raggedright\scriptsize{\parnotes}
\end{table}

\section{Conclusion}\label{sec:conclusion}
In this work, we introduced ARCADE, a novel unsupervised \ac{DL} method for network anomaly detection that automatically builds the normal traffic profile based on raw network bytes as input without human intervention for feature engineering. ARCADE is composed of a 1D-\ac{CNN} \ac{AE} that is trained exclusively on normal network traffic flows and regularized through a \ac{WGAN-GP} adversarial strategy. We experimentally demonstrated that the proposed adversarial regularization improves the performance of the \ac{AE}. The proposed adversarial regularization strategy can be applied to any \ac{AE} independently of its model architecture, and it can also be applied to other anomaly detection tasks. Once the \ac{AE} is trained on the normal network traffic using the proposed approach, ARCADE can effectively detect unseen network traffic flows from attacks and malware.  
Our results suggest that even considering only two initial packets of a network flow as input, ARCADE can detect most of the malicious traffic with nearly 100\% F1-score, except for HTTP DoS and DDoS, where 68.70\% and 66.61\% F1-scores were obtained. While considering five packets, ARCADE achieved 91.95\% and 93.19\% F1-scores for HTTP DoS and DDoS, respectively. Experiments show that the proposed \ac{AE} model is 20 times smaller than baselines and still presents significant improvement in accuracy, model complexity, and detection speed.

{
    \small
    \bibliographystyle{IEEEtranN}
    \bibliography{main} 

% Generated by IEEEtranN.bst, version: 1.14 (2015/08/26)
\begin{thebibliography}{38}
\providecommand{\natexlab}[1]{#1}
\providecommand{\url}[1]{#1}
\csname url@samestyle\endcsname
\providecommand{\newblock}{\relax}
\providecommand{\bibinfo}[2]{#2}
\providecommand{\BIBentrySTDinterwordspacing}{\spaceskip=0pt\relax}
\providecommand{\BIBentryALTinterwordstretchfactor}{4}
\providecommand{\BIBentryALTinterwordspacing}{\spaceskip=\fontdimen2\font plus
\BIBentryALTinterwordstretchfactor\fontdimen3\font minus
  \fontdimen4\font\relax}
\providecommand{\BIBforeignlanguage}[2]{{%
\expandafter\ifx\csname l@#1\endcsname\relax
\typeout{** WARNING: IEEEtranN.bst: No hyphenation pattern has been}%
\typeout{** loaded for the language `#1'. Using the pattern for}%
\typeout{** the default language instead.}%
\else
\language=\csname l@#1\endcsname
\fi
#2}}
\providecommand{\BIBdecl}{\relax}
\BIBdecl

\bibitem[Cisco(2020)]{cisco2020cisco}
U.~Cisco, ``Cisco annual internet report (2018--2023) white paper,''
  \emph{Cisco: San Jose, CA, USA}, 2020.

\bibitem[Liao et~al.(2013)Liao, Lin, Lin, and Tung]{liao2013intrusion}
H.-J. Liao, C.-H.~R. Lin, Y.-C. Lin, and K.-Y. Tung, ``Intrusion detection
  system: A comprehensive review,'' \emph{Journal of Network and Computer
  Applications}, vol.~36, no.~1, pp. 16--24, 2013.

\bibitem[Silva et~al.(2022)Silva, de~Oliveira, Medeiros, Andreoni~Lopez, and
  Mattos]{silva2022statistical}
J.~V.~V. Silva, N.~R. de~Oliveira, D.~S. Medeiros, M.~Andreoni~Lopez, and D.~M.
  Mattos, ``A statistical analysis of intrinsic bias of network security
  datasets for training machine learning mechanisms,'' \emph{Annals of
  Telecommunications}, pp. 1--17, 2022.

\bibitem[Ahmad et~al.(2021)Ahmad, Shahid~Khan, Wai~Shiang, Abdullah, and
  Ahmad]{ahmad2021network}
Z.~Ahmad, A.~Shahid~Khan, C.~Wai~Shiang, J.~Abdullah, and F.~Ahmad, ``Network
  intrusion detection system: A systematic study of machine learning and deep
  learning approaches,'' \emph{Transactions on Emerging Telecommunications
  Technologies}, vol.~32, no.~1, p. e4150, 2021.

\bibitem[Truong-Huu et~al.(2020)Truong-Huu, Dheenadhayalan, Pratim~Kundu,
  Ramnath, Liao, Teo, and Praveen~Kadiyala]{truong2020empirical}
T.~Truong-Huu, N.~Dheenadhayalan, P.~Pratim~Kundu, V.~Ramnath, J.~Liao, S.~G.
  Teo, and S.~Praveen~Kadiyala, ``An empirical study on unsupervised network
  anomaly detection using generative adversarial networks,'' in
  \emph{Proceedings of the 1st ACM Workshop on Security and Privacy on
  Artificial Intelligence}, 2020, pp. 20--29.

\bibitem[Hwang et~al.(2020)Hwang, Peng, Huang, Lin, and
  Nguyen]{hwang2020unsupervised}
R.-H. Hwang, M.-C. Peng, C.-W. Huang, P.-C. Lin, and V.-L. Nguyen, ``An
  unsupervised deep learning model for early network traffic anomaly
  detection,'' \emph{IEEE Access}, vol.~8, pp. 30\,387--30\,399, 2020.

\bibitem[Rudolph et~al.(2021)Rudolph, Wandt, and Rosenhahn]{rudolph2021same}
M.~Rudolph, B.~Wandt, and B.~Rosenhahn, ``Same same but {differNet}:
  Semi-supervised defect detection with normalizing flows,'' in
  \emph{Proceedings of the IEEE/CVF Winter Conference on Applications of
  Computer Vision}, 2021, pp. 1907--1916.

\bibitem[Gong et~al.(2019)Gong, Liu, Le, Saha, Mansour, Venkatesh, and
  Hengel]{gong2019memorizing}
D.~Gong, L.~Liu, V.~Le, B.~Saha, M.~R. Mansour, S.~Venkatesh, and A.~v.~d.
  Hengel, ``Memorizing normality to detect anomaly: Memory-augmented deep
  autoencoder for unsupervised anomaly detection,'' in \emph{Proceedings of the
  IEEE/CVF International Conference on Computer Vision}, 2019, pp. 1705--1714.

\bibitem[Zhai et~al.(2016)Zhai, Cheng, Lu, and Zhang]{zhai2016deep}
S.~Zhai, Y.~Cheng, W.~Lu, and Z.~Zhang, ``Deep structured energy based models
  for anomaly detection,'' in \emph{International conference on machine
  learning}.\hskip 1em plus 0.5em minus 0.4em\relax PMLR, 2016, pp. 1100--1109.

\bibitem[Radford et~al.(2015)Radford, Metz, and
  Chintala]{radford2015unsupervised}
A.~Radford, L.~Metz, and S.~Chintala, ``Unsupervised representation learning
  with deep convolutional generative adversarial networks,'' \emph{arXiv
  preprint arXiv:1511.06434}, 2015.

\bibitem[Goodfellow et~al.(2014)Goodfellow, Pouget-Abadie, Mirza, Xu,
  Warde-Farley, Ozair, Courville, and Bengio]{goodfellow2014generative}
I.~Goodfellow, J.~Pouget-Abadie, M.~Mirza, B.~Xu, D.~Warde-Farley, S.~Ozair,
  A.~Courville, and Y.~Bengio, ``Generative adversarial nets,'' \emph{Advances
  in neural information processing systems}, vol.~27, 2014.

\bibitem[Arjovsky et~al.(2017)Arjovsky, Chintala, and
  Bottou]{arjovsky2017wasserstein}
M.~Arjovsky, S.~Chintala, and L.~Bottou, ``Wasserstein generative adversarial
  networks,'' in \emph{International conference on machine learning}.\hskip 1em
  plus 0.5em minus 0.4em\relax PMLR, 2017, pp. 214--223.

\bibitem[Gulrajani et~al.(2017)Gulrajani, Ahmed, Arjovsky, Dumoulin, and
  Courville]{gulrajani2017improved}
I.~Gulrajani, F.~Ahmed, M.~Arjovsky, V.~Dumoulin, and A.~C. Courville,
  ``Improved training of wasserstein gans,'' \emph{Advances in neural
  information processing systems}, vol.~30, 2017.

\bibitem[Vu et~al.(2017)Vu, Bui, and Nguyen]{vu2017deep}
L.~Vu, C.~T. Bui, and Q.~U. Nguyen, ``A deep learning based method for handling
  imbalanced problem in network traffic classification,'' in \emph{Proceedings
  of the 8th international symposium on information and communication
  technology}, 2017, pp. 333--339.

\bibitem[Doriguzzi-Corin et~al.(2020)Doriguzzi-Corin, Millar, Scott-Hayward,
  Martinez-del Rincon, and Siracusa]{doriguzzi2020lucid}
R.~Doriguzzi-Corin, S.~Millar, S.~Scott-Hayward, J.~Martinez-del Rincon, and
  D.~Siracusa, ``Lucid: A practical, lightweight deep learning solution for
  ddos attack detection,'' \emph{Transactions on Network and Service
  Management}, vol.~17, no.~2, pp. 876--889, 2020.

\bibitem[Wang et~al.(2017{\natexlab{a}})Wang, Sheng, Wang, Zeng, Ye, Huang, and
  Zhu]{wang2017hast}
W.~Wang, Y.~Sheng, J.~Wang, X.~Zeng, X.~Ye, Y.~Huang, and M.~Zhu, ``Hast-ids:
  Learning hierarchical spatial-temporal features using deep neural networks to
  improve intrusion detection,'' \emph{IEEE Access}, vol.~6, pp. 1792--1806,
  2017.

\bibitem[Wang et~al.(2017{\natexlab{b}})Wang, Zhu, Zeng, Ye, and
  Sheng]{wang2017malware}
W.~Wang, M.~Zhu, X.~Zeng, X.~Ye, and Y.~Sheng, ``Malware traffic classification
  using convolutional neural network for representation learning,'' in
  \emph{International Conference on Information Networking}.\hskip 1em plus
  0.5em minus 0.4em\relax IEEE, 2017, pp. 712--717.

\bibitem[Yu et~al.(2017)Yu, Long, and Cai]{yu2017network}
Y.~Yu, J.~Long, and Z.~Cai, ``Network intrusion detection through stacking
  dilated convolutional autoencoders,'' \emph{Security and Communication
  Networks}, vol. 2017, 2017.

\bibitem[Wang et~al.(2017{\natexlab{c}})Wang, Zhu, Wang, Zeng, and
  Yang]{wang2017end}
W.~Wang, M.~Zhu, J.~Wang, X.~Zeng, and Z.~Yang, ``End-to-end encrypted traffic
  classification with one-dimensional convolution neural networks,'' in
  \emph{International Conference on Intelligence and Security
  Informatics}.\hskip 1em plus 0.5em minus 0.4em\relax IEEE, 2017, pp. 43--48.

\bibitem[Aceto et~al.(2019)Aceto, Ciuonzo, Montieri, and
  Pescap{\'e}]{aceto2019mobile}
G.~Aceto, D.~Ciuonzo, A.~Montieri, and A.~Pescap{\'e}, ``Mobile encrypted
  traffic classification using deep learning: Experimental evaluation, lessons
  learned, and challenges,'' \emph{Transactions on Network and Service
  Management}, vol.~16, no.~2, pp. 445--458, 2019.

\bibitem[Lotfollahi et~al.(2020)Lotfollahi, Jafari~Siavoshani, Shirali
  Hossein~Zade, and Saberian]{lotfollahi2020deep}
M.~Lotfollahi, M.~Jafari~Siavoshani, R.~Shirali Hossein~Zade, and M.~Saberian,
  ``Deep packet: A novel approach for encrypted traffic classification using
  deep learning,'' \emph{Soft Computing}, vol.~24, no.~3, pp. 1999--2012, 2020.

\bibitem[Ahmad et~al.(2022)Ahmad, Truscan, Vain, and Porres]{ahmad2022early}
T.~Ahmad, D.~Truscan, J.~Vain, and I.~Porres, ``Early detection of network
  attacks using deep learning,'' \emph{arXiv preprint arXiv:2201.11628}, 2022.

\bibitem[Bergmann et~al.(2018)Bergmann, L{\"o}we, Fauser, Sattlegger, and
  Steger]{bergmann2018improving}
P.~Bergmann, S.~L{\"o}we, M.~Fauser, D.~Sattlegger, and C.~Steger, ``Improving
  unsupervised defect segmentation by applying structural similarity to
  autoencoders,'' \emph{arXiv preprint arXiv:1807.02011}, 2018.

\bibitem[Wang et~al.(2004)Wang, Bovik, Sheikh, and Simoncelli]{wang2004image}
Z.~Wang, A.~C. Bovik, H.~R. Sheikh, and E.~P. Simoncelli, ``Image quality
  assessment: from error visibility to structural similarity,''
  \emph{Transactions on image processing}, vol.~13, no.~4, pp. 600--612, 2004.

\bibitem[Pang et~al.(2021)Pang, Shen, Cao, and Hengel]{pang2021deep}
G.~Pang, C.~Shen, L.~Cao, and A.~V.~D. Hengel, ``Deep learning for anomaly
  detection: A review,'' \emph{ACM Computing Surveys}, vol.~54, no.~2, pp.
  1--38, 2021.

\bibitem[Schlegl et~al.(2017)Schlegl, Seeb{\"o}ck, Waldstein, Schmidt-Erfurth,
  and Langs]{schlegl2017unsupervised}
T.~Schlegl, P.~Seeb{\"o}ck, S.~M. Waldstein, U.~Schmidt-Erfurth, and G.~Langs,
  ``Unsupervised anomaly detection with generative adversarial networks to
  guide marker discovery,'' in \emph{International conference on information
  processing in medical imaging}.\hskip 1em plus 0.5em minus 0.4em\relax
  Springer, 2017, pp. 146--157.

\bibitem[Zenati et~al.(2018)Zenati, Foo, Lecouat, Manek, and
  Chandrasekhar]{zenati2018efficient}
H.~Zenati, C.~S. Foo, B.~Lecouat, G.~Manek, and V.~R. Chandrasekhar,
  ``Efficient gan-based anomaly detection,'' \emph{arXiv:1802.06222}, 2018.

\bibitem[Akcay et~al.(2018)Akcay, Atapour-Abarghouei, and
  Breckon]{akcay2018ganomaly}
S.~Akcay, A.~Atapour-Abarghouei, and T.~P. Breckon, ``Ganomaly: Semi-supervised
  anomaly detection via adversarial training,'' in \emph{Asian conference on
  computer vision}.\hskip 1em plus 0.5em minus 0.4em\relax Springer, 2018, pp.
  622--637.

\bibitem[Deng et~al.(2009)Deng, Dong, Socher, Li, Li, and
  Fei-Fei]{deng2009imagenet}
J.~Deng, W.~Dong, R.~Socher, L.-J. Li, K.~Li, and L.~Fei-Fei, ``Imagenet: A
  large-scale hierarchical image database,'' in \emph{International conference
  on computer vision and pattern recognition}.\hskip 1em plus 0.5em minus
  0.4em\relax IEEE, 2009, pp. 248--255.

\bibitem[Xiao et~al.(2021)Xiao, Yan, and Amit]{xiao2021we}
Z.~Xiao, Q.~Yan, and Y.~Amit, ``Do we really need to learn representations from
  in-domain data for outlier detection?'' \emph{arXiv preprint
  arXiv:2105.09270}, 2021.

\bibitem[Bergman et~al.(2020)Bergman, Cohen, and Hoshen]{bergman2020deep}
L.~Bergman, N.~Cohen, and Y.~Hoshen, ``Deep nearest neighbor anomaly
  detection,'' \emph{arXiv preprint arXiv:2002.10445}, 2020.

\bibitem[Reiss et~al.(2021)Reiss, Cohen, Bergman, and Hoshen]{reiss2021panda}
T.~Reiss, N.~Cohen, L.~Bergman, and Y.~Hoshen, ``Panda: Adapting pretrained
  features for anomaly detection and segmentation,'' in \emph{Proceedings of
  the IEEE/CVF Conference on Computer Vision and Pattern Recognition}, 2021,
  pp. 2806--2814.

\bibitem[Shiravi et~al.(2012)Shiravi, Shiravi, Tavallaee, and
  Ghorbani]{shiravi2012toward}
A.~Shiravi, H.~Shiravi, M.~Tavallaee, and A.~A. Ghorbani, ``Toward developing a
  systematic approach to generate benchmark datasets for intrusion detection,''
  \emph{computers \& security}, vol.~31, no.~3, pp. 357--374, 2012.

\bibitem[McDermott et~al.(2018)McDermott, Majdani, and
  Petrovski]{mcdermott2018botnet}
C.~D. McDermott, F.~Majdani, and A.~V. Petrovski, ``Botnet detection in the
  internet of things using deep learning approaches,'' in \emph{International
  joint conference on neural networks}.\hskip 1em plus 0.5em minus 0.4em\relax
  IEEE, 2018, pp. 1--8.

\bibitem[Sch{\"o}lkopf et~al.(2001)Sch{\"o}lkopf, Platt, Shawe-Taylor, Smola,
  and Williamson]{scholkopf2001estimating}
B.~Sch{\"o}lkopf, J.~C. Platt, J.~Shawe-Taylor, A.~J. Smola, and R.~C.
  Williamson, ``Estimating the support of a high-dimensional distribution,''
  \emph{Neural computation}, vol.~13, no.~7, pp. 1443--1471, 2001.

\bibitem[Liu et~al.(2008)Liu, Ting, and Zhou]{liu2008isolation}
F.~T. Liu, K.~M. Ting, and Z.-H. Zhou, ``Isolation forest,'' in
  \emph{International conference on data mining}.\hskip 1em plus 0.5em minus
  0.4em\relax IEEE, 2008, pp. 413--422.

\bibitem[Kastryulin et~al.(2019)Kastryulin, Zakirov, and Prokopenko]{piq}
\BIBentryALTinterwordspacing
S.~Kastryulin, D.~Zakirov, and D.~Prokopenko, ``{PyTorch Image Quality}:
  Metrics and measure for image quality assessment,'' 2019, open-source
  software available at https://github.com/photosynthesis-team/piq. [Online].
  Available: \url{https://github.com/photosynthesis-team/piq}
\BIBentrySTDinterwordspacing

\bibitem[Kingma and Ba(2014)]{kingma2014adam}
D.~P. Kingma and J.~Ba, ``Adam: A method for stochastic optimization,''
  \emph{arXiv preprint arXiv:1412.6980}, 2014.

\end{thebibliography}
}

\section{Biography Section} 
\vspace{-33pt}
\begin{IEEEbiography} [{\includegraphics[width=1in,height=1.25in,clip,keepaspectratio]{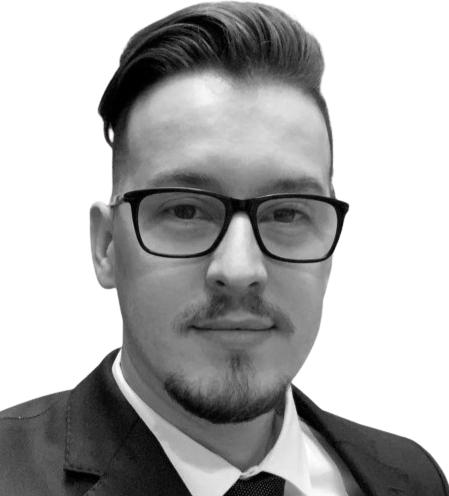}}]{Willian Tessaro Lunardi} is a Senior Research Scientist at the Secure Systems Research Centre, Technology Innovation Institute, Abu Dhabi, UAE. He has a Ph.D. in computer science from the University of Luxembourg. His main area of research is machine learning and combinatorial optimization. He is currently working on machine learning for network security, physical layer security, and jamming detection. He has published over 25 research papers in international scientific journals, conferences, and book chapters.
\end{IEEEbiography}

\vspace{-33pt}
\begin{IEEEbiography} [{\includegraphics[width=1in,height=1.25in,clip,keepaspectratio]{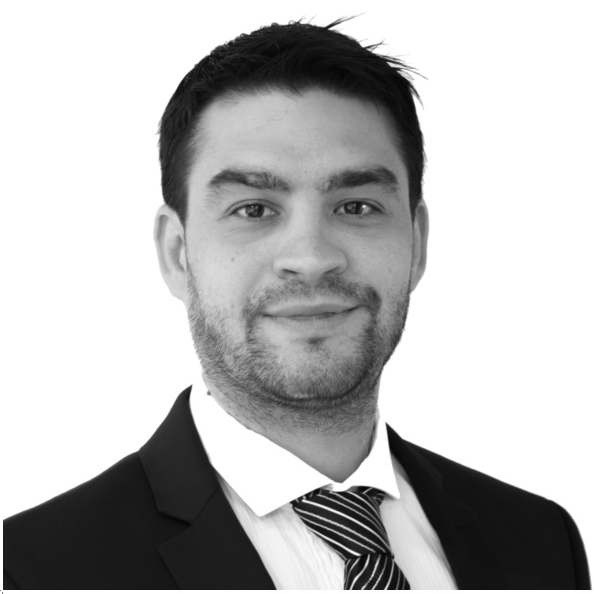}}]{Martin Andreoni Lopez} is a Network Security Researcher at the \textit{Secure System Research Center} of the \textit{Technology Innovation Institute} in Abu Dhabi, United Arab Emirates. He was a Researcher at Samsung R\&D Institute Brazil. He graduated as an Electronic Engineer from the \textit{Universidad Nacional de San Juan} (UNSJ), Argentina, in 2011. Master in Electrical Engineering from the Federal University of Rio de Janeiro (COPPE / UFRJ) in 2014. Doctor from the Federal University of Rio de Janeiro (COPPE / UFRJ) in the Teleinformatics and Automation Group (GTA) and by \textit{Sorbonne Université} in the Phare team of \textit{Laboratoire d'Informatique Paris VI} (LIP6), France, in 2018. He has co-authored several publications and patents in security, virtualization, traffic analysis, and Big Data.
\end{IEEEbiography}

\vspace{-33pt}
\begin{IEEEbiography} [{\includegraphics[width=1in,height=1.25in,clip,keepaspectratio]{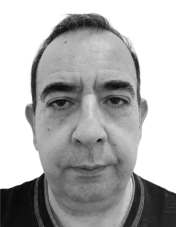}}]{Jean-Pierre Giacalone} is  Vice President of Secure Communications Engineering at Secure Systems Research Centre, Technology Innovation Institute, Abu Dhabi, UAE. He is responsible for researching secure communications, focusing on improving the resilience of cyber-physical and autonomous systems. He has worked as an expert in software architecture for advanced driving assistance systems at Renault and as principal engineer and architect within Intel's mobile systems technologies group. He has an engineering degree from the from the École nationale supérieure d'électrotechnique, d'électronique, d'informatique, d'hydraulique et des télécommunications (ENSEEIHT) in Toulouse, France. He holds 19 patents and has co-authored 15 research papers accepted for publication in international journals and conference proceedings.
\end{IEEEbiography}

\newpage
\vfill
\appendix
\begin{table}[!h]
\setlength{\tabcolsep}{4pt}\centering\scriptsize	
\caption{Encoder, decoder, and critic architecture.}\label{tab:architecture}
    \begin{tabularx}{\columnwidth}{@{}>{\hsize=3.0cm}XTTT@{}} 
        \toprule
        \textbf{Layer} & \textbf{K, S, P}$^1$ & \textbf{Output} & \textbf{Parameters} \\
        \midrule 
        \multicolumn{4}{c}{\textbf{Encoder}} \\
        \midrule
        Input & ---  & $1 \times 200$ & \\ 
        Convolution & 4, 2, 1  & $16 \times 100$ & 64 \\ 
        Batch Normalization & --- & ---  & 32  \\
        \ac{Leaky ReLU} & --- & ---  & ---  \\
        Convolution & 4, 2, 1  & $32 \times 50$ & 2,048 \\ 
        Batch Normalization & --- & ---  & 64 \\
        \ac{Leaky ReLU} & --- & ---  & ---  \\
        Convolution & 4, 2, 1  & $64 \times 25$ & 8,192 \\ 
        Batch Normalization & --- & ---  & 128 \\
        \ac{Leaky ReLU} & --- & ---  & --- \\
        Linear & ---  & $50$ & 80,000 \\ 
        \midrule
        Mean &   &  & 90,528 \\ 
        \midrule
        \multicolumn{4}{c}{\textbf{Decoder}} \\
        \midrule
        Input & ---  & $50$ & \\ 
        Linear & ---  & $64 \times 25$ & 80,000 \\ 
        Batch Normalization & --- & ---  & 3,200  \\
        \ac{ReLU} & --- & ---  & ---  \\
        Transposed Convolution & 4, 2, 1  & $32 \times 50$ & 8,192 \\ 
        Batch Normalization & --- & ---  & 64  \\
        \ac{ReLU} & --- & ---  & ---  \\
        Transposed Convolution & 4, 2, 1  & $16 \times 100$ & 2,048 \\ 
        Batch Normalization & --- & ---  & 32 \\
        \ac{ReLU} & --- & ---  & ---  \\
        Transposed Convolution & 4, 2, 1 & $1 \times 200$ & 64 \\ 
        Sigmoid & --- & ---  & ---  \\
        \midrule
        Mean &   &  & 93,600 \\ 
        \midrule
        \multicolumn{4}{c}{\textbf{Critic}} \\
        \midrule
        Input & ---  & $1 \times 200$ & \\ 
        Convolution & 4, 2, 1  & $16 \times 100$ & 64 \\ 
        Layer Normalization & --- & ---  & 3,200  \\
        \ac{Leaky ReLU} & --- & ---  & ---  \\
        Convolution & 4, 2, 1  & $32 \times 50$ & 2,048 \\ 
        Layer Normalization & --- & ---  & 3,200  \\
        \ac{Leaky ReLU} & --- & ---  & ---  \\
        Convolution & 4, 2, 1  & $64 \times 25$ & 8,192 \\ 
        Layer Normalization & --- & ---  & 3,200 \\
        \ac{Leaky ReLU} & --- & ---  & ---  \\
        Linear & ---  & $50$ & 80,000 \\ 
        Layer Normalization & --- & ---  & 100 \\
        \ac{Leaky ReLU} & --- & ---  & ---  \\
        Linear & ---  & $1$ & 50 \\  
        \midrule
        Mean &   &  & 100,105 \\  
        \bottomrule
    \end{tabularx}
    \raggedright
    \scriptsize
    $^{1}$Kernel, Stride, Padding
\end{table}   

\begin{table}[!h]
\setlength{\tabcolsep}{4pt}\centering\scriptsize	
\caption{Specifications of considered environments for detection speed experiments.}\label{tab:spec}
    \begin{tabularx}{\columnwidth}{@{}>{\hsize=2.25cm}XT>{\hsize=0.8cm}TTT@{}}  %>{\hsize=3.0cm}
        \toprule
        \textbf{Device} & \textbf{OS}$^1$ & \textbf{Mem.}$^2$ & \textbf{CPU} & \textbf{GPU} \\
        \midrule
        Raspberry Pi 4B         & Pi OS Lite 32-bit   & 4GB   & Cortex-A72                & --- \\ 
        UP Xtreme WHLI7         & Ubuntu 20.10 64-bit & 16GB  & Intel Core i7-8665UE      & --- \\ 
        Jetson Xavier NX        & Jetson Linux R32.7.1 64-bit & 8GB   & Carmel ARM                & NVIDIA Volta with 384 CUDA cores and 48 tensor cores  \\    
        Desktop 1               & Ubuntu 20.10 64-bit & 128GB & AMD Ryzen Threadripper 3970X   & --- \\ 
        Desktop 2               & Ubuntu 20.10 64-bit & 128GB & AMD Ryzen Threadripper 3970X   & NVIDIA RTX 3090 with 10,496 CUDA cores and 328 tensor cores \\ 
        \bottomrule
    \end{tabularx}
\end{table}    

\end{document}